\documentclass[10pt,twocolumn,letterpaper]{article}

\usepackage[pagenumbers]{cvpr} %

\graphicspath{{figs/}}
\usepackage{multirow}
\usepackage{array}
\usepackage{float}
\floatstyle{ruled}
\newfloat{algorithm}{t}{loa}
\floatname{algorithm}{Algorithm}

\usepackage{listings}
\lstdefinestyle{prompt}{basicstyle=\ttfamily\scriptsize,breaklines=true,
  breakindent=0pt,columns=fullflexible,keepspaces=true,frame=single,framesep=3pt,
  xleftmargin=2pt,xrightmargin=2pt,aboveskip=4pt,belowskip=2pt,upquote=true}

\usepackage[most]{tcolorbox}
\tcbset{chat/.style={boxrule=0.3pt,arc=3pt,left=5pt,right=5pt,top=2.5pt,bottom=2.5pt,
  boxsep=1pt,fontupper=\footnotesize,width=0.86\columnwidth}}
\newtcolorbox{userbubble}{chat,colback=blue!8,colframe=blue!35!black!40,flush right}
\newtcolorbox{modelbubble}{chat,colback=gray!12,colframe=gray!45,flush left}
\newtcolorbox{ubmbubble}{chat,colback=red!6,colframe=red!35!black!35,flush left}
\newtcolorbox{refbubble}{chat,colback=green!8,colframe=green!40!black!35,flush left}

\definecolor{cvprblue}{rgb}{0.21,0.49,0.74}
\usepackage[pagebackref,breaklinks,colorlinks,allcolors=cvprblue]{hyperref}
\usepackage[capitalize]{cleveref}   %
\crefname{algorithm}{Algorithm}{Algorithms}
\Crefname{algorithm}{Algorithm}{Algorithms}

\usepackage{xspace}
\newcommand{\ours}{RepConTUR\xspace}
\newcommand{\refseg}{RES\xspace}
\newcommand{\gseg}{GSEG\xspace}

\newcommand{\RRG}{RRG\xspace}
\newcommand{\VQA}{VQA\xspace}

\def\confName{CVPR}
\def\confYear{2027}

\newif\ifmergeappendix
\ifdefined\bodyonly\mergeappendixfalse\else\mergeappendixtrue\fi

\title{RepConTUR: A Report-Conditioned Volumetric Multimodal Model for
       Abdominal Tumor Segmentation}

\author{
Seunghwan Lee$^{1}$ \quad Sangjoon Park$^{2}$ \quad Jongmin Lee$^{1}$ \quad Yoon Jo Kim$^{1}$\\
Wonyoung Cho$^{1}$ \quad Jin Sung Kim$^{1,2}$ \quad Han Joo Chae$^{1,}$\thanks{Corresponding author.}\\[0.6em]
$^{1}$Oncosoft Inc., Seoul, South Korea\\
$^{2}$Department of Radiation Oncology, Yonsei University College of Medicine, Seoul, South Korea
}

\begin{document}
\maketitle
\begin{abstract}
Tumor delineation is required for radiotherapy planning, yet automatic segmentation is limited by
errors, predominantly false positives, that distort dose and increase editing time. The radiology
report accompanies the scan and localizes disease at the suborgan level, but existing text-promptable models
condition on short referring phrases, and most operate in 2D. We present RepConTUR, a single native-3D early-fusion multimodal model
that conditions volumetric abdominal tumor segmentation on the narrative report. 
We extend AbdomenAtlas 3.0 to a suborgan-level promptable segmentation dataset and generate report-based VQA pairs,
yielding four report-derived tasks on the same annotations: referring segmentation
(RES), report-grounded segmentation (GSEG), report generation (RRG), and visual question
answering (VQA). We train RepConTUR, comprising a
3D CT encoder, an early-fusion language model, and a native-3D segmentation module, jointly on the
four tasks. Across liver, kidney and
pancreas, RepConTUR attains the highest Dice similarity coefficient (DSC), instance-wise precision, and
detection F1 among the text-promptable baselines, and adding the report raises mean precision while
leaving recall comparable. Against two vision-only specialists, nnU-Net and the report-supervised R-Super, RepConTUR attains the highest value on every in-domain metric for liver and pancreas
and the highest mean DSC, recall, precision, and detection F1 over the four out-of-distribution public cohorts.
Ablations indicate that conditioning on the narrative report raises detection F1 from 0.499 to 0.528 within the same model, that joint GSEG training raises RES F1 from 0.450 to 0.482, and that the native-3D segmentation module raises GSEG F1 from 0.139 to 0.528 over its 2D counterpart. Code is available at \href{https://github.com/Oncosoft-Research/RepConTUR}{github.com/Oncosoft-Research/RepConTUR}.
\end{abstract}

\section{Introduction}
\label{sec:intro}

Tumor delineation is a required step in radiotherapy planning, but manual
contouring is labor-intensive and shows large inter-observer variability~\cite{vinod2016uncertainties,
segedin2016uncertainties, ginn2023timesavings}. Unlike diagnostic reading, radiotherapy target
contouring routinely combines the diagnostic scan with its radiology report, as the report
contains rich information about the tumor, including its anatomical location, size and
stage~\cite{icru83_2010, bird2015multimodality, eisenhauer2009recist, amin2017ajcc8}. This makes the report a natural conditioning signal for automatic tumor
segmentation.

Vision-only deep learning models remain the standard approach to tumor segmentation, and their
accuracy is reported mainly by the volumetric Dice similarity coefficient (DSC), which
underrepresents instance-wise recall and precision~\cite{maierhein2024metrics}. On the LiTS
liver-tumor benchmark, the top method reports a tumor Dice of $0.702$, yet its instance-wise precision
is only $0.148$~\cite{bilic2023lits}. In
radiotherapy, instance-wise recall and precision are as clinically important as DSC. False positives in particular distort the
dose distribution and raise normal-tissue exposure and toxicity risk~\cite{poel2022outliers}, and these errors increase the editing time~\cite{vaassen2020apl}.
A separate line of work uses the radiology report as training-time supervision to improve tumor
detection and segmentation. Report-derived tumor count and size penalize inconsistent predictions
through auxiliary losses~\cite{bassi2025rsuper}, reported location provides weak
labels~\cite{guo2024boosting}, and the report yields pseudo-labels~\cite{bosma2023reportguided}.
However, these models still segment from the image alone at inference time~\cite{ye2024sgseg}, so the report cannot directly inform the prediction.

Text-guided segmentation models have achieved strong results on referring segmentation (\refseg)
and grounded segmentation (\gseg)~\cite{yang2022lavt, lai2024lisa, rasheed2024glamm, zou2023seem,
li2022glip}. Some works adopt a dual-encoder design that encodes the language prompt and the image
in separate branches and fuses them at the decoder~\cite{zhao2025foundation,
voxtell2024}. On the other hand, multimodal large language model (MLLM)-based segmentation models fuse image and
text early and reason over both jointly within the language model~\cite{lai2024lisa, wu2026unibiomed}. Yet both are designed for
short target descriptions --- class names or referring phrases --- and do not exploit long-form
textual context. BiomedParse v2, for
example, conditions on an organ-level prompt (e.g., ``tumor in liver CT''), which cannot indicate a
detailed intra-organ region and may not capture suborgan-level information~\cite{zhao2025foundation, zhao2025boltzformer}.
A report instead specifies suborgan location, which may help suppress false positives elsewhere in
the same organ~\cite{couinaud1957, lirads2018}. Furthermore, existing MLLM-based medical segmentation models are also mostly
2D~\cite{wu2026unibiomed}, limiting their abilities to interpret report information that refers to 3D context~\cite{isensee2021nnunet, cicek20163dunet, mlynarski2019longrange}.

Multitask learning can improve a target task when the tasks are related, although the gain depends on
task affinity~\cite{caruana1997multitask, ruder2017overview, standley2020which}. Unified vision--language
models train localization and language tasks in one network~\cite{zhang2022glipv2, wang2022ofa}, and an
MLLM-based segmentation model is a natural fit for such joint training. In the medical domain, UniBioMed trains segmentation jointly with \VQA and report generation and
reports improved segmentation Dice~\cite{wu2026unibiomed}. The report can therefore also
serve as a joint-training signal.

\begin{figure*}[t]
  \centering
  \begin{minipage}{0.80\textwidth}\centering
  \begin{minipage}[c]{0.51\linewidth}\centering
    \begin{subfigure}{\linewidth}\centering
      \includegraphics[width=\linewidth]{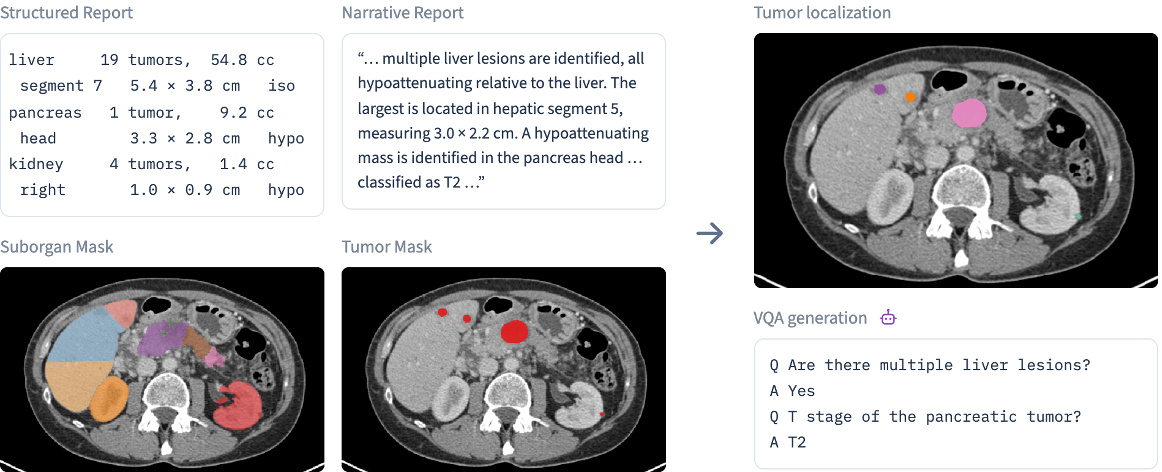}
      \caption{}
      \label{fig:overview-data}
    \end{subfigure}
    \\[6pt]
    \begin{subfigure}{\linewidth}\centering
      \includegraphics[width=\linewidth]{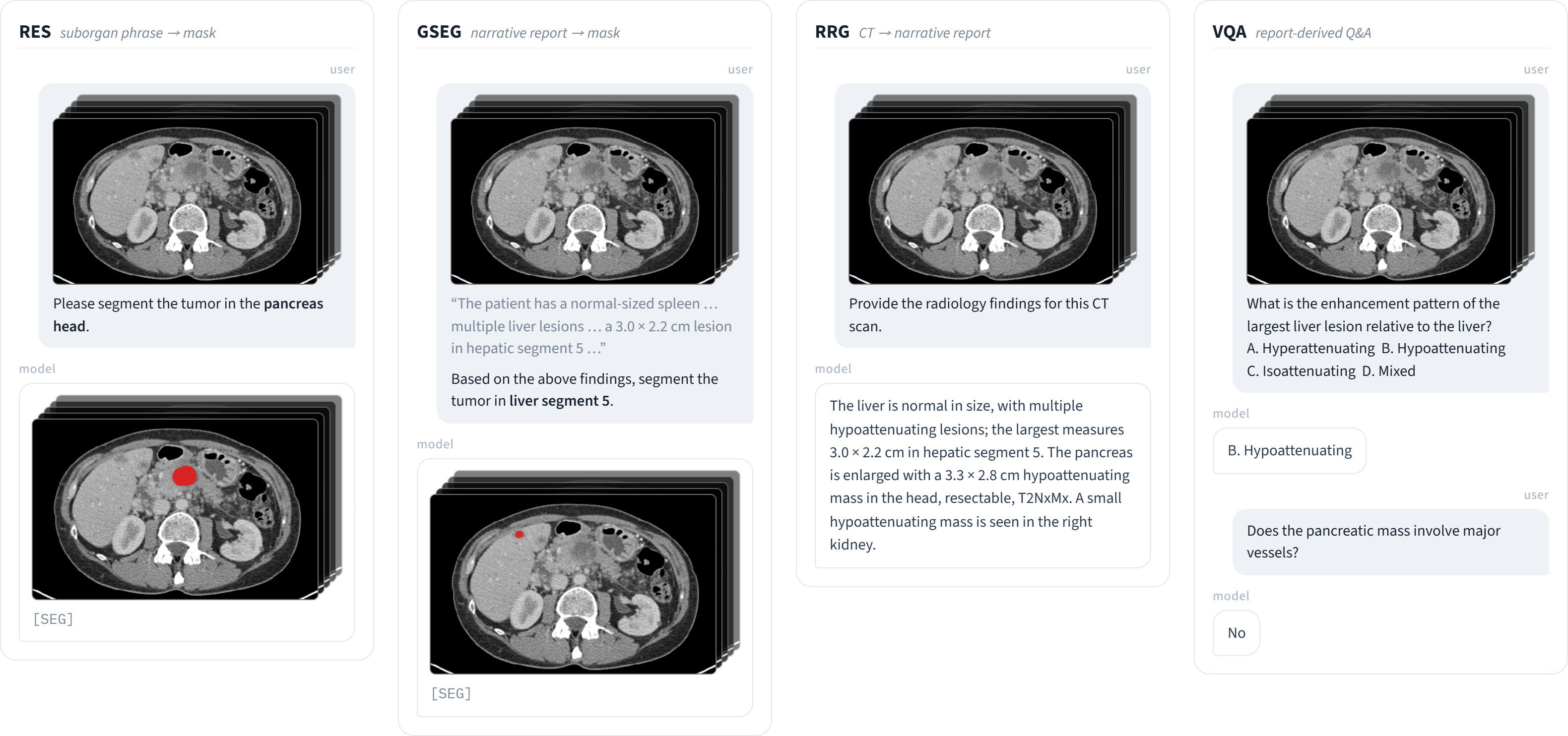}
      \caption{}
      \label{fig:overview-tasks}
    \end{subfigure}
  \end{minipage}
  \hfill
  \begin{minipage}[c]{0.47\linewidth}\centering
    \begin{subfigure}{\linewidth}\centering
      \includegraphics[width=\linewidth]{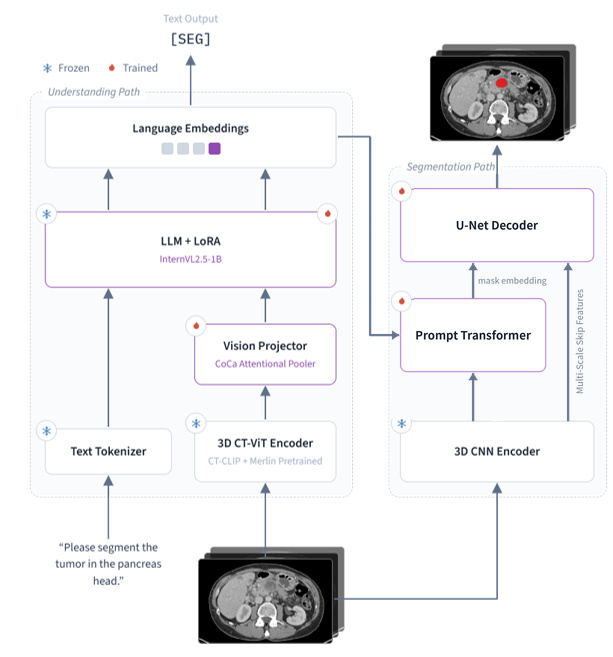}
      \caption{}
      \label{fig:overview-model}
    \end{subfigure}
  \end{minipage}
  \end{minipage}
  \caption{\textbf{Overview of \ours.} \textbf{(a)} Report-driven dataset construction from
  AbdomenAtlas 3.0~\cite{bassi2025radgpt}. Report matching decomposes each organ tumor mask into
  suborgan-localized instances by intersection-over-component, yielding the \refseg and \gseg targets;
  the reports additionally yield verified \VQA pairs. \textbf{(b)} The four tasks trained jointly on
  the shared backbone, namely \refseg, \gseg, \RRG and \VQA. \textbf{(c)} A single native-3D
  early-fusion model. A 3D CT encoder produces visual tokens fused with the text prompt from the first
  LLM layer; a \texttt{[SEG]} token drives a 3D segmentation module to a volumetric mask, while the
  same backbone answers \VQA and generates \RRG.}
  \label{fig:overview}
\end{figure*}

We propose \ours, an MLLM-based 3D report-conditioned tumor segmentation
model that grounds segmentation on the report to improve both segmentation accuracy and instance-wise
detection (\cref{fig:overview}). Our contributions are threefold:
\begin{itemize}\setlength{\itemsep}{1pt}
  \item \textbf{Report-conditioned segmentation.} We condition segmentation on a suborgan-level
        referring phrase and an optional free-form narrative report, rather than a short class name
        alone. To the best of our knowledge, \ours is the first 3D referring/grounded segmentation
        model to condition on such report context.
        We introduce suborgan-level mask and prompt design aligned to narrative free-form radiology reports, and implement it
        for abdominal tumors.
  \item \textbf{A native 3D MLLM for report-conditioned abdominal tumor segmentation.}
        We build on prior MLLM architectures and extend them with native-3D vision modules. The MLLM vision encoder and segmentation module are all 3D, 
        and visual tokens are processed jointly with the report text from the first LLM layer, so vision and text are reasoned over jointly.
  \item \textbf{Report-derived multitask dataset and multitask learning.} From AbdomenAtlas 3.0~\cite{bassi2025radgpt}
        tumor masks and their reports we construct a unified four-task dataset. Report-matched suborgan localization produces the \refseg and \gseg targets, and the report is converted into radiology report generation (\RRG) supervision and LLM-verified \VQA pairs. We jointly train all four tasks on a single backbone. Cumulative task ablations show that report-grounded segmentation training improves \refseg performance, while adding report generation and \VQA further increases detection F1.
\end{itemize}

\section{Related work}
\label{sec:related}

\subsection{Reports and tumor segmentation}
One line of work uses the report as a training signal. R-Super converts report-derived
tumor count and size into voxel-wise losses~\cite{bassi2025rsuper}, Guo \etal use report text as
weak supervision to improve detection~\cite{guo2024boosting}, Bosma \etal derive pseudo-labels from
the report~\cite{bosma2023reportguided}, and SGSeg uses text only during training and drops it at
inference~\cite{ye2024sgseg}. 

These methods all use the report only at training time. In contrast, we condition 3D tumor
segmentation on the narrative report at inference.

\subsection{Text-promptable medical segmentation}
These methods share a common pipeline. A separate text encoder maps a short query to an embedding
that is fused with vision features, usually late at the decoder, to produce a text-conditioned mask.
The two families differ only in the query. Open-vocabulary methods use a class name or anatomical
term as the query (SegVol~\cite{du2024segvol}, SAT~\cite{zhao2025sat},
CLIP-Universal~\cite{liu2023clip}, BiomedParse v2~\cite{zhao2025foundation, zhao2025boltzformer}),
while a parallel line uses a referring or descriptive phrase (LViT~\cite{li2023lvit},
Ariadne's Thread~\cite{zhong2023ariadne}, RecLMIS~\cite{huang2025reclmis},
CausalCLIPSeg~\cite{chen2024causalclipseg}, LSMS~\cite{ouyang2024lsms}, VoxTell~\cite{voxtell2024}) or attribute labels such as
tumor count and severity (TGANet~\cite{tomar2022tganet}, Bi-VLGM~\cite{chen2024bivlgm}).

These methods primarily target class-level or phrase-level queries, leaving their effectiveness with
full narrative reports less established. We investigate report conditioning by jointly processing
volumetric visual tokens and report text in an MLLM before mask prediction.

\subsection{MLLM-based segmentation}
This family splits into two modules. An MLLM processes text and image jointly to produce an embedding aligned to the image, and
a separate segmentation module combines that embedding with fine-grained
backbone features to produce the mask. LISA introduced this design for natural
images~\cite{lai2024lisa} and later works extended it~\cite{rasheed2024glamm, yuan2025sa2va,
xu2023ullava, wei2024lasagna}.
A medical branch adapts this design to biomedical images, usually in 2D. UniBioMed~\cite{wu2026unibiomed}
couples an MLLM with a SAM-style segmentation module for grounded biomedical interpretation.
MedPLIB~\cite{huang2024medplib} drives a separate SAM-Med mask decoder from a \texttt{[SEG]}
embedding, with the pixel-grounding path routed through an in-LLM mixture-of-experts. 

\ours inherits this aligned embedding plus separate segmentation module design, but extends the
2D medical setting to full 3D and broadens the conditioning from a short referring phrase to the
narrative report.

\subsection{Multitask learning}
In the medical domain, multitask training most often pairs segmentation with a diagnostic
objective. Joint segmentation-and-classification improves segmentation over single-task
training~\cite{zhou2021multitask}, and universal models unify detection, segmentation and diagnosis
across organs (CancerUniT~\cite{chen2023cancerunit}, CLIP-Universal~\cite{liu2023clip}). A newer
line instead couples segmentation with language tasks such as diagnostic text
generation~\cite{li2025raddiagsegm}, \VQA~\cite{huang2024medplib} and report
generation~\cite{gu2025interactive}. UniBioMed trains segmentation jointly with \VQA and report generation
and reports improved segmentation Dice~\cite{wu2026unibiomed}.

We adopt this report-based multitask scheme and build a report-derived
multitask dataset from AbdomenAtlas 3.0. From report-aligned tumor localization we derive two suborgan-level tumor segmentation tasks (\refseg and \gseg), and from the report itself we derive \RRG and \VQA. We jointly train all four on full-3D CT volumes and condition segmentation on the narrative report.

\section{Report-guided multitask dataset}
\label{sec:method}
\label{sec:data}
We build the report-guided multitask dataset from AbdomenAtlas 3.0~\cite{bassi2025radgpt}, which
provides $9{,}262$ abdominal CT volumes with tumor masks for liver, kidney and pancreas, and organ
masks that include suborgan structures such as liver Couinaud segments. Each volume also carries a
synthetically generated radiology report from the RadGPT construction pipeline~\cite{bassi2025radgpt}.
A structured report is first derived deterministically from the masks, listing per suborgan the tumor
count, size, location, and attenuation. A large language model then rewrites the structured report
into a narrative free-text report in radiologist style. The narrative may not enumerate every tumor.
For multifocal disease it often reports the largest tumor with its suborgan and an aggregate
tumor count rather than each tumor's location. Across the $9{,}262$ volumes, the narrative reports have a mean sequence length of $225$ tokens and a
maximum of $1{,}163$ tokens under the Clinical-Longformer tokenizer~\cite{li2023clinicallongformer}.

We exclude volumes without tumors in any of the three organs to match the clinical setting of radiotherapy, where planning is performed only for tumor-positive patients. Of the $9{,}262$ volumes, $5{,}464$ are
tumor-free and removed, leaving $3{,}798$ tumor-bearing volumes for training and evaluation. Positives
are unevenly distributed across organs, with $1{,}508$ liver-positive, $2{,}222$ kidney-positive and
$1{,}113$ pancreas-positive volumes, namely $39.7\%$, $58.5\%$ and $29.3\%$. Every organ is therefore trained with both
tumor-present and tumor-absent volumes.

\textbf{Report-guided tumor localization.} The original tumor masks are at organ level, so we
decompose them into per-suborgan instances using the structured report. We extract the connected
components (CC) of the tumor mask. For each component $c$ and suborgan mask $s$ we measure their
intersection over component,

\begin{equation}
\mathrm{IoC}(c,s)=\frac{|c\cap s|}{|c|},
\end{equation}

and associate $c$ with every suborgan whose IoC is nonzero, since one component can span more than one
suborgan. The structured report describes the tumors at two levels of location, and we assign components
differently for the two. Most findings name the suborgan and state how many tumors it contains, whereas
others give only an organ-level finding without a suborgan. For a suborgan named by the report, we rank
its overlapping components by IoC and keep the highest-ranked ones up to the tumor count stated for that
suborgan. For a tumor the report places only at the organ level, the component is assigned to every suborgan whose IoC exceeds a fixed threshold and to its maximum-IoC suborgan. A component that overlaps no suborgan is instead assigned by spatial proximity. 
Because a suborgan can hold several tumors, we merge all components assigned to it into a single mask by voxel-wise union.

\textbf{VQA generation.} We generate up to $20$ question--answer pairs per volume from the structured
and narrative reports in a generation stage followed by a verification stage. The structured report
supplies the exact per-tumor facts queried, namely suborgan location, size, and attenuation, while the
narrative supplies tumor staging and clinical interpretation such as resectability. In the generation stage, a large language model
writes pairs in three answer formats, namely binary yes/no, short answer, and multiple choice with
enumerated options. The pairs span tumor presence, tumor count, suborgan location, tumor size,
enhancement pattern relative to the organ, and organ-level findings such as organ enlargement. Tumor
size is queried as an interval rather than an exact measurement, since an exact
diameter varies with the measurer while a coarse band is reproducible, using the bands $<\!1$,
$1$--$2$, $2$--$3$, $3$--$5$, $5$--$10$ and $>\!10$~cm. ``No'' questions are generated by perturbing a stated finding into a
plausible but incorrect variation such as a different suborgan, side, or stage. In the verification
stage, a separate language model checks each pair against the source structured and narrative reports
and corrects or discards any pair that is unsupported by the reports, not answerable from the image alone, malformed in
its answer, insufficiently localized, or whose negative is not a plausible variation. Across the $3{,}798$ tumor-bearing volumes, verification retains $71{,}783$ pairs, $18.9$ per
volume on average, comprising $48{,}451$ yes/no, $13{,}433$ short-answer and $9{,}899$ multiple-choice.
Within the yes/no pairs, $31{,}449$ ($64.9\%$) are affirmative and $17{,}002$ ($35.1\%$) negative.

\textbf{Multitask composition.} We train \ours jointly on four objectives, namely \refseg and \gseg
for segmentation, \RRG for radiology report generation and \VQA for visual question answering.
\refseg conditions on a suborgan-level referring phrase drawn from a pool of phrasal variants that
indicate the tumor in a suborgan, for example ``tumor in hepatic segment 7''. At training time one phrase is sampled from this pool and inserted into a randomly sampled instruction template,
such as ``Please segment the tumor in \{phrase\}.'' or ``Please delineate the GTV
in \{phrase\}.''. \refseg also includes a whole-organ target that aggregates every tumor in an organ,
named by an organ-level phrase such as ``total liver GTV'' or ``all tumors in the kidney''. \gseg reuses
the same target mask and referring phrase but prepends the full narrative report to the instruction,
such as ``\{report\}\,\ldots\ Based on the above findings, please segment the tumor in
\{phrase\}.''. The comparison therefore measures the additional benefit of report context beyond the
target phrase, with the image and requested target held fixed.

\RRG generates the report from the CT, and \VQA
answers the report-derived questions. The two language tasks supply language supervision to the shared
backbone that the segmentation tasks reuse.

\section{3D MLLM segmentation model}
\label{sec:model}

\textbf{Early-fusion 3D MLLM.} We adopt the CT-ViT of CT-CLIP~\cite{hamamci2024ctrate} as the 3D visual
encoder. It factorizes attention into a spatial transformer that
attends within each axial plane followed by a temporal transformer that attends across planes along
the through-slice axis, producing a 3D token grid. This encodes the whole volume in 3D at lower cost
than full 3D attention. We vary the original implementation in its input, resampling volumes to a $(1.5,1.5,3.0)$~mm spacing and a $240\times240\times140$ grid.
We pretrain the encoder with a CLIP objective, aligning each CT volume embedding with a
Clinical-Longformer~\cite{li2023clinicallongformer} embedding of its report. The original CT-ViT is
pretrained only on chest CT from CT-RATE~\cite{hamamci2024ctrate}, so we add abdominal CT image--report
pairs from Merlin~\cite{blankemeier2024merlin} to generalize the encoder from chest to abdominal
radiological semantics, which our downstream abdominal tasks require. After pretraining we keep the
frozen image branch as a feature extractor.

Following CoCa~\cite{yu2022coca}, an
attentional pooler with 256 learnable queries cross-attends to the encoder tokens to
produce 256 fixed-length visual tokens, projected to the LLM hidden dimension.
We adopt a InternVL2.5-$1$B~\cite{chen2024internvl} language model as the backbone, initialized from
the medically pretrained UniBioMed~\cite{wu2026unibiomed} weights. The LLM is frozen and adapted with
LoRA~\cite{hu2022lora}. We introduce a \texttt{[SEG]} token for requesting the segmentation~\cite{lai2024lisa}.

\textbf{3D segmentation module.} We adopt VoxTell~\cite{voxtell2024} as the 3D segmentation module,
initialized from its pretrained weights. It pairs an nnU-Net residual encoder~\cite{isensee2021nnunet}
with a text-prompted decoder that consists of a prompt transformer and a U-Net decoder. We keep the
encoder frozen and train the decoder.

We construct this query from the language model. The hidden state of the \texttt{[SEG]} token is mapped
by a two-layer projection to a $2048$-dimensional query embedding. Because the MLLM fuses image and text
early, this embedding jointly encodes the volume and the input text. The prompt transformer
cross-attends from it to the encoder features to form a mask embedding, which the U-Net decoder fuses
at every scale to produce the full-resolution volumetric mask.

\section{Experiments and results}
\label{sec:exp}

\subsection{Setup}
\label{sec:exp-setup}
We evaluate abdominal tumor delineation on liver, kidney and pancreas across two streams, plus design
ablations.

\textbf{Text-promptable comparison.} The first stream evaluates text-promptable segmentation under a
single referring prompt, where \ours is driven either by \refseg mode or \gseg mode. We compare it against three
text-promptable baselines, namely the 2D MLLM-based model
UniBioMed~\cite{wu2026unibiomed}, the 2D late-fusion model BiomedParse v2~\cite{zhao2025foundation, zhao2025boltzformer},
and the 3D late-fusion model VoxTell~\cite{voxtell2024}. We finetune each baseline on our dataset
with its own finetuning recipe, restricted to the tasks it supports, namely BiomedParse v2 on \refseg,
VoxTell on \refseg and \gseg, and UniBioMed on all four tasks. This stream is evaluated on the
AbdomenAtlas 3.0 test split.

\textbf{Report-guided organ-wise segmentation.} The second stream evaluates report-guided organ-wise
segmentation against two vision-only specialists, the nnU-Net specialist~\cite{isensee2021nnunet}
trained to segment liver, kidney and pancreas tumors jointly as separate classes, and
R-Super~\cite{bassi2025rsuper}, a segmentation model supervised by radiology reports during training;
both use no report at inference. To simulate the radiotherapy
setting where a CT scan and its report are available together, we run a simple agentic inference
pipeline in which an LLM parses the report into the tumor-bearing suborgans, \ours grounds and segments
each suborgan under \refseg or \gseg, and the per-suborgan masks are combined by union into an
organ-wise prediction. When the narrative report aggregates findings at the organ level and does not
localize the tumor to a specific suborgan, the LLM cannot name a suborgan target, and we prompt \ours with a
single whole-organ tumor phrase for that organ. Because neither specialist can ground individual suborgans, the report enters them only as
an organ-level restriction that zeroes any organ the report marks tumor-free. This stream is
evaluated on an in-domain test set and four external out-of-distribution (OOD) sets,
namely HCC-TACE-Seg for primary hepatocellular carcinoma~\cite{hcctaceseg2023, hcctacesegdata},
Colorectal-Liver-Metastases for colorectal liver metastases~\cite{crlm2024, crlmdata}, CPTAC-PDA for pancreatic
ductal adenocarcinoma~\cite{cptacpda, cptacpdaannot}, and CPTAC-CCRCC for clear cell renal cell
carcinoma~\cite{cptacccrcc, cptacccrccannot}. These OOD sets have no radiology reports, so for each case we synthesize
one with the same construction pipeline used for AbdomenAtlas 3.0~\cite{bassi2025radgpt} and use it
both to drive the agentic inference pipeline and as the \gseg grounding text.

\textbf{Ablations.} We ablate two design choices on the test split. The first evaluates successive
replacements of the 2D visual and segmentation modules with their 3D counterparts, namely a 2D/2D
reference, a 3D/2D variant, and the full 3D/3D model (\cref{tab:des}). The second adds the four
training tasks cumulatively, from \refseg alone to the full multitask model (\cref{tab:mt}).

\textbf{Evaluation metrics.} We assess our model with DSC, instance-wise
Recall (R) and Precision (P), and their harmonic mean F1. Tumor instances are 3D connected components,
and a predicted tumor matches a ground-truth tumor by IoU$\ge0.10$, following medical detection
practice~\cite{baumgartner2021nndetection, saha2024picai}.

\subsection{Comparison with text-promptable baselines}
\label{sec:exp-main}
\begin{table}[t]\centering\footnotesize\setlength{\tabcolsep}{5pt}
\caption{Comparison with the text-promptable baselines on test split.
Rs$=$\refseg, Gs$=$\gseg. DSC: Dice, R/P:
instance-wise Recall/Precision, F1: harmonic mean.
\textbf{Bold}: highest per column per section.}\label{tab:main}
\resizebox{\columnwidth}{!}{%
\begin{tabular}{l cc cc cc cc}
\toprule
 & \multicolumn{2}{c}{DSC} & \multicolumn{2}{c}{R} & \multicolumn{2}{c}{P} & \multicolumn{2}{c}{F1} \\
\cmidrule(lr){2-3}\cmidrule(lr){4-5}\cmidrule(lr){6-7}\cmidrule(lr){8-9}
Model & Rs & Gs & Rs & Gs & Rs & Gs & Rs & Gs \\
\midrule
\multicolumn{9}{l}{\textbf{Liver}} \\
VoxTell        & 0.406 & 0.339 & 0.455 & 0.387 & 0.019 & 0.010 & 0.037 & 0.020 \\
UniBioMed      & 0.370 & 0.391 & 0.531 & 0.554 & 0.081 & 0.086 & 0.141 & 0.149 \\
BiomedParse v2 & 0.341 & ---   & \textbf{0.693} & --- & 0.082 & --- & 0.147 & --- \\
\ours          & \textbf{0.522} & \textbf{0.542} & 0.616 & \textbf{0.611} & \textbf{0.369} & \textbf{0.424} & \textbf{0.461} & \textbf{0.500} \\
\midrule
\multicolumn{9}{l}{\textbf{Pancreas}} \\
VoxTell        & 0.270 & 0.221 & 0.430 & 0.336 & 0.099 & 0.078 & 0.161 & 0.127 \\
UniBioMed      & 0.205 & 0.215 & 0.399 & 0.413 & 0.032 & 0.039 & 0.060 & 0.072 \\
BiomedParse v2 & 0.305 & ---   & \textbf{0.711} & --- & 0.057 & --- & 0.105 & --- \\
\ours          & \textbf{0.386} & \textbf{0.399} & 0.624 & \textbf{0.644} & \textbf{0.336} & \textbf{0.376} & \textbf{0.437} & \textbf{0.475} \\
\midrule
\multicolumn{9}{l}{\textbf{Kidney}} \\
VoxTell        & 0.470 & 0.391 & 0.477 & 0.369 & 0.025 & 0.020 & 0.047 & 0.039 \\
UniBioMed      & 0.450 & 0.456 & 0.619 & 0.612 & 0.070 & 0.069 & 0.126 & 0.123 \\
BiomedParse v2 & 0.469 & ---   & \textbf{0.836} & --- & 0.075 & --- & 0.138 & --- \\
\ours          & \textbf{0.621} & \textbf{0.627} & 0.781 & \textbf{0.775} & \textbf{0.487} & \textbf{0.501} & \textbf{0.600} & \textbf{0.609} \\
\midrule
\multicolumn{9}{l}{\textbf{Mean}} \\
VoxTell        & 0.382 & 0.317 & 0.454 & 0.364 & 0.048 & 0.036 & 0.082 & 0.062 \\
UniBioMed      & 0.342 & 0.354 & 0.516 & 0.526 & 0.061 & 0.065 & 0.109 & 0.115 \\
BiomedParse v2 & 0.372 & ---   & \textbf{0.747} & --- & 0.071 & --- & 0.130 & --- \\
\ours          & \textbf{0.510} & \textbf{0.523} & 0.674 & \textbf{0.677} & \textbf{0.397} & \textbf{0.433} & \textbf{0.499} & \textbf{0.528} \\
\bottomrule
\end{tabular}}
\end{table}
On the test split (\cref{tab:main}), \ours attains the highest DSC, instance-wise precision, and detection F1
in every organ section for both \refseg and \gseg, and the highest recall against every baseline except
BiomedParse v2. Averaged over the three organs, \ours
precision is $0.397$ in \refseg and $0.433$ in \gseg, versus $\le0.071$ for every baseline. The sole
exception is recall, where BiomedParse v2 is higher in every organ, with mean recall
$0.747$ against \ours \refseg $0.674$, but its precision is substantially lower, so \ours attains a higher detection F1
by a substantial margin, mean $0.499$ against $0.130$. Between the two conditioning modes
of \ours, prepending the report raises mean precision from \refseg $0.397$ to \gseg $0.433$,
while mean recall is comparable at $0.674$ and $0.677$, so the report-conditioning effect is concentrated in precision.
The baselines respond differently to report conditioning. VoxTell's macro F1 decreases from $0.082$
in \refseg to $0.062$ in \gseg, whereas UniBioMed and \ours improve from $0.109$ to $0.115$ and from
$0.499$ to $0.528$, respectively. The MLLM-based models benefit from the additional report text, while
VoxTell, a late-fusion model, does not.

\begin{figure*}[t]\centering
  \setlength{\tabcolsep}{1pt}\renewcommand{\arraystretch}{0.5}
  \newcommand{\qimg}[1]{$\vcenter{\hbox{\includegraphics[width=0.15\textwidth]{#1}}}$}
  \newcommand{\qlab}[2]{$\vcenter{\hbox{\scriptsize\shortstack[c]{\textbf{#1}\\[1pt]``#2''}}}$}
  \begin{tabular}{@{}c ccccc@{}}
     & \scriptsize GT & \scriptsize\textbf{\ours} & \scriptsize VoxTell & \scriptsize UniBioMed & \scriptsize BiomedParse v2 \\
    \qlab{Liver}{tumor in segment 4 of the liver} &
      \qimg{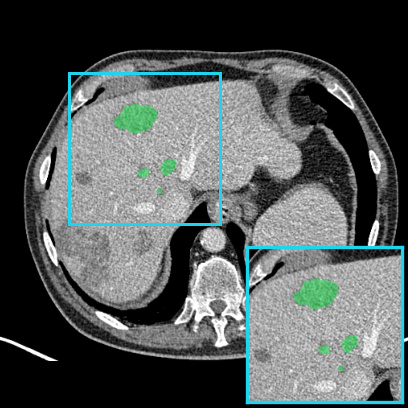} & \qimg{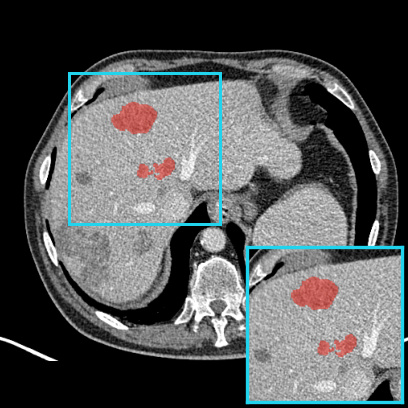} & \qimg{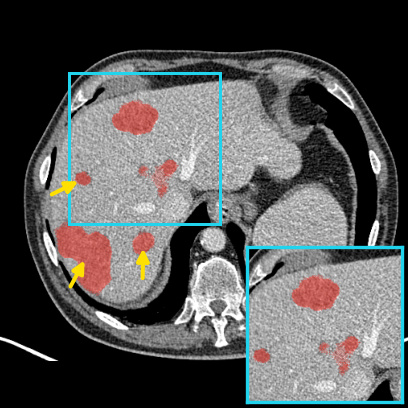} & \qimg{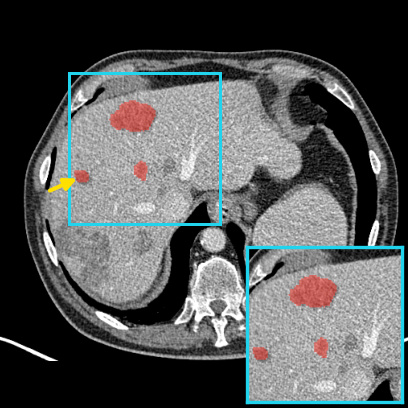} & \qimg{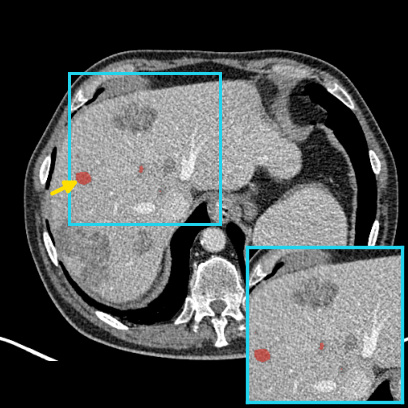} \\
    \qlab{Kidney}{tumor in the kidney} &
      \qimg{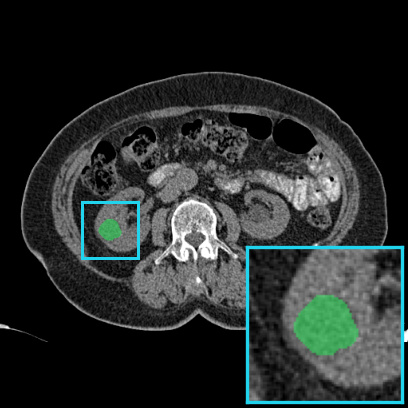} & \qimg{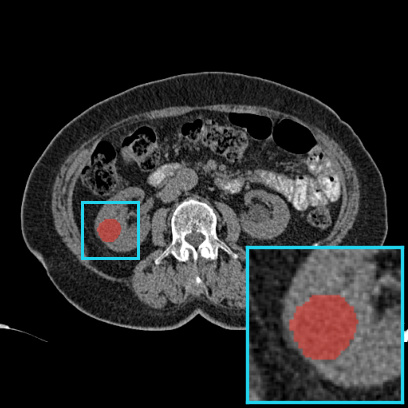} & \qimg{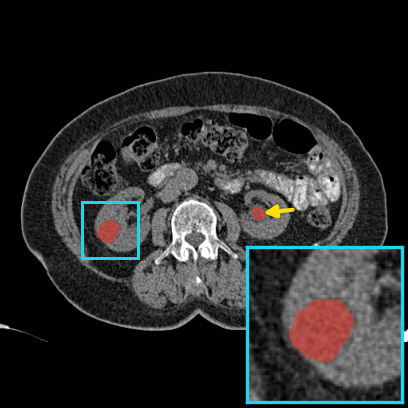} & \qimg{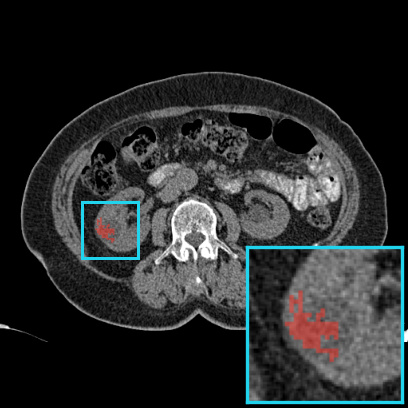} & \qimg{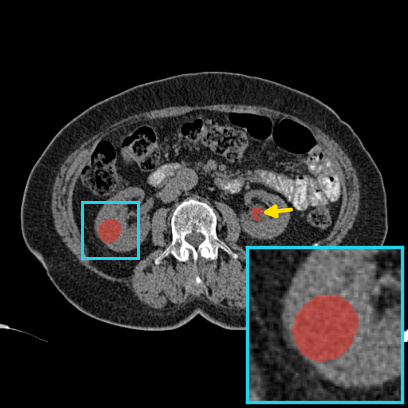} \\
    \qlab{Pancreas}{tumor in the pancreas body} &
      \qimg{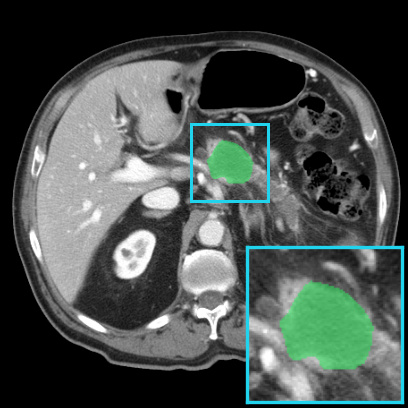} & \qimg{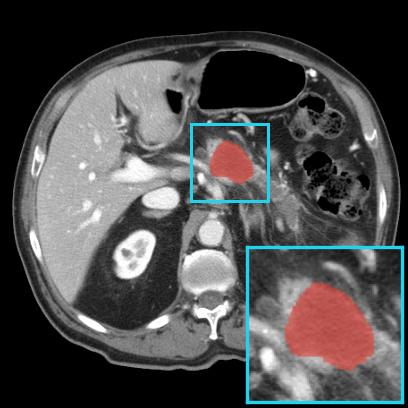} & \qimg{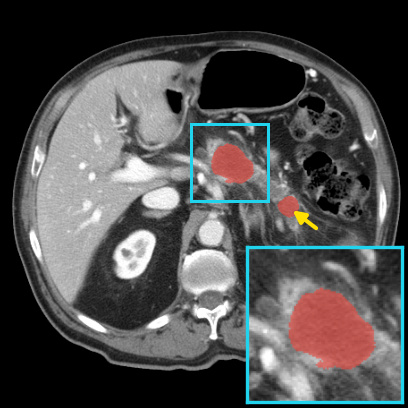} & \qimg{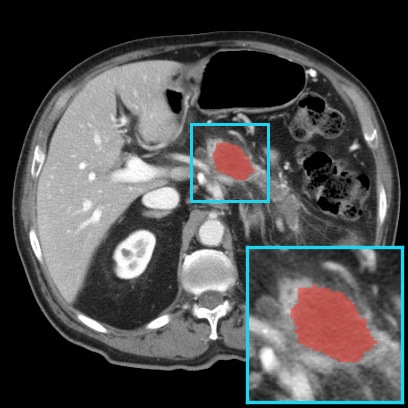} & \qimg{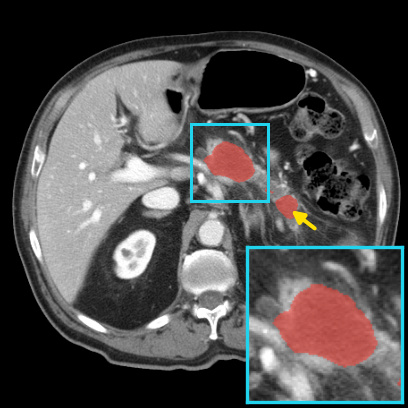} \\
  \end{tabular}
  \caption{Qualitative segmentation on the test split, same referring prompt per row, one axial slice.
  Cyan box marks the referred tumor, magnified in the lower-right inset; yellow arrows mark components
  outside it. GT green, prediction red. \ours confines its prediction to the referred tumor, whereas
  the baselines add components elsewhere on the slice, namely other real but non-referred tumors for
  the multifocal liver and pancreas cases and the tumor-free contralateral kidney for the kidney case.
  These raise the false-positive count and are consistent with the low baseline precision in
  \cref{tab:main}.}\label{fig:qual}
\end{figure*}

\begin{figure*}[t]\centering
  \footnotesize
  \begin{subfigure}{0.24\linewidth}\includegraphics[width=\linewidth]{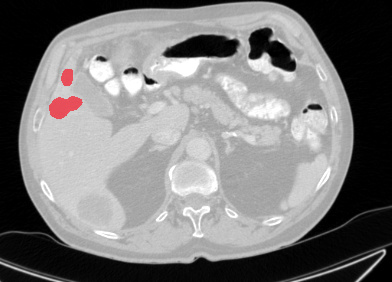}\caption*{Segment 5}\end{subfigure}\hfill
  \begin{subfigure}{0.24\linewidth}\includegraphics[width=\linewidth]{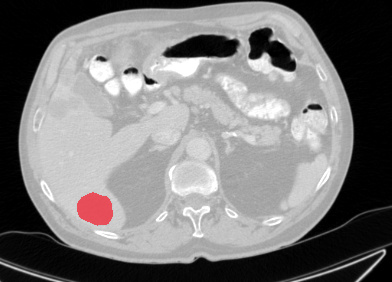}\caption*{Segment 6}\end{subfigure}\hfill
  \begin{subfigure}{0.24\linewidth}\includegraphics[width=\linewidth]{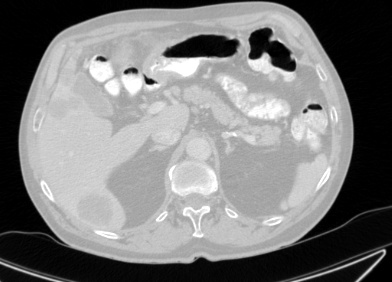}\caption*{Segment 3}\end{subfigure}\hfill
  \begin{subfigure}{0.24\linewidth}\includegraphics[width=\linewidth]{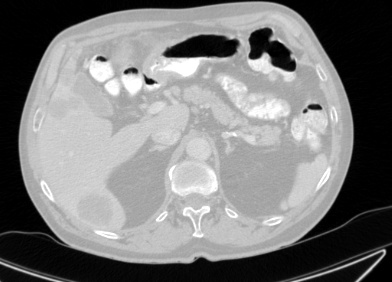}\caption*{Kidney}\end{subfigure}
  \caption{Prompt conditioning for a multifocal liver case, one axial slice. Switching the prompt
  between segments 5 and 6 moves the mask from anterior to posterior; querying segment 3, whose tumor
  lies on other slices, or the tumor-free kidney yields an empty mask, so the prompt controls both
  location and presence.}\label{fig:prompt}
\end{figure*}

\textbf{Qualitative results and prompt conditioning} Across the three organs,
\ours confines its prediction to the referred tumor, whereas the baselines
add false positives on the same slice (\cref{fig:qual}). The liver and pancreas cases each contain
multiple tumors, so these false positives are not spurious activations but genuine tumors elsewhere in
the same organ that the referring prompt did not single out. Under the
pancreatic-body prompt the baselines additionally delineate a separate tumor in the tail, and under
the segment-4 liver prompt they add further hepatic tumors. In the kidney they instead segment the
tumor-free contralateral organ. Both behaviours are localization failures relative to the referring
prompt and are consistent with the low baseline precision in \cref{tab:main}. \Cref{fig:prompt}
shows that the model does not segment indiscriminately. Changing the suborgan in the prompt
moves the predicted mask, and querying a tumor-free organ produces an empty output, so the prompt
determines both the location and the presence of the segmentation.

\subsection{Ablation studies}
\label{sec:exp-des}
\begin{table}[t]\centering\footnotesize\setlength{\tabcolsep}{4pt}
\caption{Design ablation replacing the 2D visual and segmentation modules with 3D counterparts,
3-organ macro means. Rows are (vision / segmentation) module; 2D/2D is UniBioMed, 3D/2D replaces only
the visual encoder. Metrics and abbreviations as in \cref{tab:main}. \textbf{Bold}: highest per column.}\label{tab:des}
\resizebox{\columnwidth}{!}{%
\begin{tabular}{l cc cc cc cc}
\toprule
 & \multicolumn{2}{c}{DSC} & \multicolumn{2}{c}{R} & \multicolumn{2}{c}{P} & \multicolumn{2}{c}{F1} \\
\cmidrule(lr){2-3}\cmidrule(lr){4-5}\cmidrule(lr){6-7}\cmidrule(lr){8-9}
Vis./Seg. & Rs & Gs & Rs & Gs & Rs & Gs & Rs & Gs \\
\midrule
2D / 2D & 0.342 & 0.354 & 0.517 & 0.526 & 0.061 & 0.065 & 0.109 & 0.115 \\
3D / 2D & 0.331 & 0.378 & 0.526 & 0.535 & 0.075 & 0.080 & 0.129 & 0.139 \\
\midrule
3D / 3D & \textbf{0.510} & \textbf{0.523} & \textbf{0.674} & \textbf{0.677} & \textbf{0.397} & \textbf{0.433} & \textbf{0.499} & \textbf{0.528} \\
\bottomrule
\end{tabular}}
\end{table}

\Cref{tab:des} shows an ablation that varies the vision encoder and the segmentation module from 2D
to 3D. The 2D/2D row pairs a 2D InternViT vision encoder~\cite{chen2024internvl} with a SAM2
segmentation module~\cite{ravi2024sam2}, initialized with UniBioMed. The 3D/2D row replaces this 2D
encoder with our 3D CT-ViT while keeping the 2D SAM2 segmentation module, and the 3D/3D row adds the
3D segmentation module to form the full \ours. Replacing the 2D visual encoder with the 3D CT-ViT
yields modest improvements in detection metrics in both modes. DSC increases in \gseg from $0.354$ to
$0.378$ but decreases in \refseg from $0.342$ to $0.331$. The larger improvement occurs when the 2D
segmentation module is replaced with the pretrained 3D module, with \gseg precision increasing from
$0.080$ to $0.433$ and detection F1 from $0.139$ to $0.528$.

\begin{table}[t]\centering\footnotesize\setlength{\tabcolsep}{4pt}
\caption{Multitask ablation, tasks added cumulatively (\refseg, +\gseg, +\RRG, +\VQA), 3-organ macro
means. A \refseg-only model has no Gs (--). Metrics and abbreviations as in \cref{tab:main}.
\textbf{Bold}: highest per column.}\label{tab:mt}
\resizebox{\columnwidth}{!}{%
\begin{tabular}{l cc cc cc cc}
\toprule
 & \multicolumn{2}{c}{DSC} & \multicolumn{2}{c}{R} & \multicolumn{2}{c}{P} & \multicolumn{2}{c}{F1} \\
\cmidrule(lr){2-3}\cmidrule(lr){4-5}\cmidrule(lr){6-7}\cmidrule(lr){8-9}
Train & Rs & Gs & Rs & Gs & Rs & Gs & Rs & Gs \\
\midrule
\refseg   & 0.481 & ---            & 0.657 & ---            & 0.345 & ---            & 0.450 & ---            \\
+\gseg    & 0.505 & \textbf{0.524} & 0.650 & 0.658          & 0.385 & 0.415          & 0.482 & 0.508          \\
+\RRG     & 0.503 & \textbf{0.524} & 0.656 & 0.655          & 0.396 & \textbf{0.435} & 0.493 & 0.523          \\
\midrule
+\VQA     & \textbf{0.510} & 0.523 & \textbf{0.674} & \textbf{0.677} & \textbf{0.397} & 0.433 & \textbf{0.499} & \textbf{0.528} \\
\bottomrule
\end{tabular}}
\end{table}
\Cref{tab:mt} adds the training tasks cumulatively and reports both inference modes together.
Starting from a \refseg-only model, adding \gseg co-training gives the largest single gain. It brings
the report-grounded \gseg mode to a strong operating point, with DSC $0.524$ and F1 $0.508$, and it
also raises the phrase-only \refseg mode that uses no report at inference, lifting DSC from $0.481$ to
$0.505$, precision from $0.345$ to $0.385$, and F1 from $0.450$ to $0.482$. This transfer to \refseg
occurs despite that mode sharing no report text at test time. Adding the two language tasks, \RRG then
\VQA, further increases detection F1, from $0.482$ to $0.493$ to $0.499$ in \refseg and from $0.508$
to $0.523$ to $0.528$ in \gseg. Other metrics do not improve monotonically, and \gseg DSC remains
stable at $0.523$--$0.524$. The observed benefit of the additional language tasks is therefore
concentrated in detection F1.

\subsection{Report-based inference versus vision-only specialists}
\label{sec:exp-report}
Following the report-guided organ-wise protocol of \cref{sec:exp-setup}, we compare the two
report-based conditions, \refseg and \gseg, against two vision-only specialists. We report
results separately for the in-domain test set (\cref{tab:specialist}) and the four
external OOD sets (\cref{tab:external}).

\begin{table}[t]\centering\footnotesize\setlength{\tabcolsep}{6pt}
\caption{Report-based inference on the in-domain test set, per-organ \refseg/\gseg vs.\ two
vision-only specialists, namely nnU-Net and the report-supervised R-Super~\cite{bassi2025rsuper}. 
nnU-Net and R-Super are restricted to the organs the agentic pipeline names,
\refseg/\gseg are suborgan-decomposed. Metrics as in \cref{tab:main}. \textbf{Bold}: highest per
column per organ.}\label{tab:specialist}
\begin{tabular*}{\columnwidth}{@{\extracolsep{\fill}}l cccc}
\toprule
Model & DSC & R & P & F1 \\
\midrule
\multicolumn{5}{l}{\textbf{Liver} ($n{=}134$)} \\
nnU-Net & 0.602 & 0.566 & 0.392 & 0.463 \\
R-Super & 0.580 & 0.517 & 0.457 & 0.485 \\
\refseg & 0.610 & \textbf{0.659} & 0.550 & \textbf{0.599} \\
\gseg   & \textbf{0.612} & 0.633 & \textbf{0.559} & 0.594 \\
\midrule
\multicolumn{5}{l}{\textbf{Pancreas} ($n{=}110$)} \\
nnU-Net & 0.367 & 0.554 & 0.297 & 0.387 \\
R-Super & 0.392 & 0.612 & 0.359 & 0.452 \\
\refseg & 0.393 & 0.619 & 0.350 & 0.447 \\
\gseg   & \textbf{0.403} & \textbf{0.626} & \textbf{0.383} & \textbf{0.475} \\
\midrule
\multicolumn{5}{l}{\textbf{Kidney} ($n{=}225$)} \\
nnU-Net & \textbf{0.664} & 0.751 & \textbf{0.590} & \textbf{0.661} \\
R-Super & 0.652 & \textbf{0.784} & 0.435 & 0.559 \\
\refseg & 0.626 & 0.695 & 0.503 & 0.584 \\
\gseg   & 0.625 & 0.689 & 0.493 & 0.575 \\
\midrule
\multicolumn{5}{l}{\textbf{Mean}} \\
nnU-Net & 0.544 & 0.624 & 0.426 & 0.504 \\
R-Super & 0.542 & 0.638 & 0.417 & 0.499 \\
\refseg & 0.543 & \textbf{0.658} & 0.468 & 0.543 \\
\gseg   & \textbf{0.547} & 0.649 & \textbf{0.478} & \textbf{0.548} \\
\bottomrule
\end{tabular*}
\end{table}
\begin{table}[t]\centering\footnotesize\setlength{\tabcolsep}{6pt}
\caption{Out-of-distribution generalization on four public tumor sets outside AbdomenAtlas 3.0,
\refseg/\gseg (suborgan-decomposed, synthetic reports) vs.\ the vision-only nnU-Net and
report-supervised R-Super~\cite{bassi2025rsuper} specialists. Sets are GT-positive only, so no
organ is suppressed. Metrics as in \cref{tab:main}. \textbf{Bold}: highest per column per
cohort.}\label{tab:external}
\begin{tabular*}{\columnwidth}{@{\extracolsep{\fill}}l cccc}
\toprule
Model & DSC & R & P & F1 \\
\midrule
\multicolumn{5}{l}{\textbf{HCC-TACE-Seg} --- HCC (liver), $n{=}130$} \\
nnU-Net & 0.650 & 0.679 & 0.134 & 0.223 \\
R-Super & 0.597 & 0.653 & 0.111 & 0.190 \\
\refseg & 0.719 & 0.689 & 0.225 & 0.340 \\
\gseg   & \textbf{0.727} & \textbf{0.694} & \textbf{0.293} & \textbf{0.412} \\
\midrule
\multicolumn{5}{l}{\textbf{CPTAC-PDA} --- PDAC (pancreas), $n{=}108$} \\
nnU-Net & 0.492 & 0.766 & 0.287 & 0.418 \\
R-Super & 0.471 & 0.793 & \textbf{0.449} & \textbf{0.573} \\
\refseg & 0.488 & 0.793 & 0.389 & 0.522 \\
\gseg   & \textbf{0.503} & \textbf{0.802} & 0.443 & 0.571 \\
\midrule
\multicolumn{5}{l}{\textbf{CPTAC-CCRCC} --- ccRCC (kidney), $n{=}108$} \\
nnU-Net & 0.804 & \textbf{0.943} & 0.240 & 0.383 \\
R-Super & 0.791 & \textbf{0.943} & 0.187 & 0.312 \\
\refseg & 0.799 & 0.927 & 0.303 & 0.457 \\
\gseg   & \textbf{0.807} & 0.927 & \textbf{0.348} & \textbf{0.506} \\
\midrule
\multicolumn{5}{l}{\textbf{Colorectal-Liver-Metastases} --- CRLM (liver), $n{=}197$} \\
nnU-Net & \textbf{0.672} & 0.731 & 0.565 & 0.637 \\
R-Super & 0.611 & \textbf{0.756} & 0.539 & 0.630 \\
\refseg & 0.625 & 0.731 & 0.643 & 0.684 \\
\gseg   & 0.626 & 0.737 & \textbf{0.644} & \textbf{0.687} \\
\midrule
\multicolumn{5}{l}{\textbf{Mean}} \\
nnU-Net & 0.655 & 0.780 & 0.307 & 0.415 \\
R-Super & 0.618 & 0.786 & 0.322 & 0.426 \\
\refseg & 0.658 & 0.785 & 0.390 & 0.501 \\
\gseg   & \textbf{0.666} & \textbf{0.790} & \textbf{0.432} & \textbf{0.544} \\
\bottomrule
\end{tabular*}
\end{table}

\paragraph{In-domain test set.}
On liver and pancreas, the two tumor types with the least conspicuous boundaries, \refseg and
\gseg improve tumor detection substantially (\cref{tab:specialist}). Detection F1 rises to $0.599$ on
liver (\refseg) and $0.475$ on pancreas (\gseg). R-Super also exceeds the vision-only nnU-Net specialist on
detection F1 for liver and pancreas, at $0.485$ against $0.463$ and $0.452$ against $0.387$. \ours improves
further with the report available at inference, clearly on liver at $0.599$ in \refseg against R-Super
$0.485$, and comparably to R-Super on pancreas at $0.447$ in \refseg and $0.475$ in \gseg against
$0.452$. Hepatic and pancreatic tumors are frequently low-contrast and blend into the surrounding
parenchyma, so the report context appears to help localize disease that vision alone tends to miss or
over-segment. On kidney, whose tumors are comparatively well-delineated in contrast-enhanced CT, the
vision-only nnU-Net specialist is already strong~\cite{heller2021kits19} at F1 $0.661$, while R-Super
falls to $0.559$ through a precision collapse to $0.435$ despite the highest kidney recall of $0.784$,
and \ours sits between at $0.575$--$0.584$. Averaged over the three organs, \refseg and \gseg exceed
both specialists on detection F1, $0.543$ and $0.548$ against nnU-Net $0.504$ and R-Super $0.499$, and
on precision; the R-Super mean is held down by kidney rather than resembling nnU-Net across organs.

\paragraph{External OOD sets.}
\Cref{tab:external} reports the four external OOD cohorts. \gseg attains the highest detection F1 in
three of four cohorts and is comparable to the report-supervised R-Super specialist on CPTAC-PDA
($0.571$ vs.\ $0.573$); it exceeds the nnU-Net specialist by $0.189$ on HCC, $0.153$ on PDA, $0.123$
on CCRCC and $0.050$ on CRLM. The margin over nnU-Net is positive on all four, including the kidney
cohort CCRCC, the one organ on which nnU-Net was higher in domain, so the advantage extends even where
vision-only segmentation was strongest in domain. The gain comes from false-positive suppression, as
the nnU-Net precision degrades out of domain to a mean $0.307$ against \gseg $0.432$, while recall
stays comparable across methods. R-Super, which uses reports for training-time supervision, shows the
same out-of-domain precision collapse as nnU-Net at a mean $0.322$, so \ours attains higher
external-cohort precision than both evaluated specialists. The advantage is not specific to \gseg. \refseg,
which uses report-derived target phrases without conditioning the segmentation model on the full
report, also exceeds the nnU-Net specialist F1 in every cohort, by $0.117$ on HCC,
$0.104$ on PDA, $0.074$ on CCRCC and $0.047$ on CRLM. The margin over the nnU-Net specialist is also larger out
of domain than in domain, with a mean detection-F1 advantage of $0.129$ for \gseg and $0.086$ for
\refseg out of domain against $0.044$ and $0.039$ in domain, consistent with report-guided grounding
generalizing under domain shift better than a vision-only specialist. The contrast between \refseg and
\gseg isolates the report-text effect, since the two share the same suborgan decomposition and differ
only in report text. F1 rises with report text on the primary cancers, by $0.072$ on HCC, $0.049$ on
PDA, $0.049$ on CCRCC, and $0.003$ on CRLM, where
DSC is instead higher for nnU-Net at $0.672$ against $0.626$. On CRLM \ours still surpasses the nnU-Net specialist on detection F1, driven by higher
precision.

\subsection{Limitations and future work}
\label{sec:limitations}

\ours targets three abdominal organs, namely
liver, kidney and pancreas, while tumors in the other body regions are excluded in this work.
Whether the approach scales to a larger set of tumor classes, spanning further abdominal organs and
other body regions, is not established, and verifying this class-scaling ability is a primary
direction for future work. 

The validation cohorts lack real reports and were grounded with synthetic ones. Although these
synthetic reports emulate the style of real narrative radiology reports, whether the same gain holds
under real clinical reports, given their stylistic variation and reporting inconsistencies, remains to
be verified. Confirming the benefit requires evaluation on datasets that pair real-world radiology
reports with their images.

The clinical value of automatic segmentation rests on the editing time it saves, which we do not measure on our outputs~\cite{vaassen2020apl}, so prospective clinical and editing-time studies remain future work.

\section{Conclusion}
\label{sec:conclusion}
We presented \ours, a native-3D multimodal model that conditions abdominal tumor segmentation on a referring phrase and, optionally, a narrative report, using report-aligned suborgan masks. A report-derived dataset supports joint training on four tasks from AbdomenAtlas 3.0, namely referring and grounded segmentation, report generation and question answering. \ours attains the highest DSC, precision and detection F1 among the text-promptable baselines, and improves mean detection F1 over the nnU-Net and R-Super specialists in-domain and across four external cohorts, although performance varies by organ. These results support report-conditioned segmentation under synthetic reports, and validation with independently authored clinical reports remains necessary.

\section*{LLM Usage Statement}
We used large language models in data construction, inference and manuscript preparation.
GPT-5.4 generated the \VQA pairs from the AbdomenAtlas 3.0 reports, and GPT-5.4-mini verified them
by correcting or discarding pairs inconsistent with the report.
OpenAI's \texttt{gpt-5.6-terra} rewrote the structured reports of the four external cohorts into
narrative reports, as detailed in the supplementary material.
The same model parses each report into tumor-bearing suborgan targets in the agentic inference
pipeline (\cref{sec:exp-setup}).
Claude (Anthropic), used through Claude Code, assisted in writing and debugging code, running
experiments, producing figures, surveying related work, organizing and verifying references, and
writing and editing the manuscript, and it provided review feedback on manuscript drafts.
OpenAI Codex was also used to verify citations and to review the manuscript.
No LLM output was used without review by the authors, who take full responsibility for all content.

\section*{Acknowledgements}
This research was supported by the AI Computing Infrastructure Enhancement (GPU Rental Support) User Support Program
funded by the Ministry of Science and ICT (MSIT), Republic of Korea.

{
    \small
    \bibliographystyle{ieeenat_fullname}
    \bibliography{main}
}

\ifmergeappendix
  \clearpage
  \raggedbottom
  \appendix
\section{Appendix}
\setcounter{table}{0}\setcounter{figure}{0}\setcounter{algorithm}{0}
\renewcommand{\thetable}{\thesection.\arabic{table}}
\renewcommand{\thefigure}{\thesection.\arabic{figure}}
\renewcommand{\thealgorithm}{\thesection.\arabic{algorithm}}

\subsection{Method and implementation details}

\subsubsection{Training and inference}
\label{app:preproc}
This section details the CT preprocessing, training and inference pipelines.

\textbf{Target construction.} The decomposition of \cref{sec:data} uses a fixed IoC threshold of $0.2$
for a component the report places only at the organ level. Spatial proximity is measured as the surface
distance from a component to a suborgan. The surface-distance rule also fills a stated slot when a
suborgan's reported tumor count exceeds its overlapping components. Across the $14{,}584$ tumor
components, the suborgan-count step accounts for $11{,}958$ and the organ-level rule handles the
remaining $2{,}618$, while IoC assigns $13{,}723$ and surface distance assigns $853$. Splitting the two
jointly, $80.4\%$ of components are suborgan-stated and assigned by IoC, $1.6\%$ suborgan-stated by
surface distance, $13.7\%$ organ-level by IoC, and $4.3\%$ organ-level by surface distance.

\textbf{CT preprocessing.} The language encoder and the segmentation module follow different pipelines.
The language encoder consumes a fixed $240\times240\times140$ volume, produced by resampling to
$(1.5,1.5,3.0)$~mm and then center-cropping or padding, clipped to $[-1000,1000]$~HU and min--max
normalized to $[-1,1]$. The segmentation module trains on $1$~mm isotropic, z-score-normalized $192^3$
patches. The segmentation volume is first padded so every dimension is at least $192$, and a $192^3$
patch is then drawn per volume by a foreground-biased crop that samples tumor-positive and
tumor-negative centers in equal proportion. Aside from the random crop location, no data augmentation
is applied.

\textbf{Losses.} Segmentation uses voxel-wise sigmoid cross-entropy and Dice loss summed $1{:}1$, and
the language tasks use standard language-model cross-entropy. The two are summed according to the task
composition of each batch.

\textbf{Optimization.} \ours trains for $20$ epochs on four NVIDIA H100 80GB GPUs in bf16 mixed
precision. Distributed training uses the DeepSpeed ZeRO-2 strategy, with a per-GPU batch of $4$ and
gradient accumulation of $8$, giving an effective batch of $128$ and a maximum sequence length of
$8192$. The optimizer is AdamW ($\beta{=}(0.9,0.999)$, weight decay $0.05$) with learning rate
$4\!\times\!10^{-5}$ and gradient norms clipped to $1$, using one epoch of linear warmup at ratio
$0.05$ followed by cosine annealing. The LoRA adapter has rank $64$ and $\alpha{=}128$.

\textbf{VQA generation.} The \VQA pairs are generated with GPT-5.4 and verified with GPT-5.4-mini,
which corrects or discards pairs inconsistent with the report.

\textbf{Inference.} At inference \ours segments the full volume by sliding-window inference with a
$192^3$ window and $50\%$ overlap, caching the language-model features of the fixed-grid volume and
prompt so they are computed once and reused across suborgan prompts, as detailed in \cref{app:compute}.

\subsubsection{Out-of-distribution sets}
\label{app:ood}

\paragraph{Cohort construction and exclusion criteria.}
The four out-of-distribution cohorts (\cref{tab:external}) are built from public TCIA collections,
namely HCC-TACE-Seg~\cite{hcctaceseg2023, hcctacesegdata}, Colorectal-Liver-Metastases~\cite{crlm2024, crlmdata},
CPTAC-PDA~\cite{cptacpda, cptacpdaannot} and CPTAC-CCRCC~\cite{cptacccrcc, cptacccrccannot}. The four cohorts span three organs and
both primary and metastatic disease, and each targets a distinct clinical scenario. HCC-TACE-Seg
contains primary hepatocellular carcinoma imaged in a transarterial-chemoembolization cohort, so it
probes delineation of primary liver tumors whose appearance is altered by locoregional therapy.
Colorectal-Liver-Metastases contains multifocal colorectal liver metastases, so it probes
sensitivity to several scattered deposits rather than to a single tumor. CPTAC-PDA contains
pancreatic ductal adenocarcinoma, an infiltrative and low-contrast primary, so it probes the hardest
abdominal target, where small tumors are the primary detection challenge. CPTAC-CCRCC contains clear
cell renal cell carcinoma and probes primary renal-tumor delineation, the organ on which the
vision-only specialist was strongest in domain. 

The evaluation unit is the CT series rather than the patient, so a patient with several eligible
contrast phases contributes several scans. The HCC-TACE-Seg CT series interleaves two to three
contrast phases, so we separate the phases by acquisition and keep every phase that contains the
tumor, yielding $130$ scans from $105$ patients. CPTAC-PDA and CPTAC-CCRCC likewise yield $108$
scans from $67$ and $47$ patients, and Colorectal-Liver-Metastases contributes one scan per patient
for $197$ scans. We exclude contours drawn on MR or PET series and discard any scan whose contours do
not rasterize to a non-empty tumor mask on the CT grid.

\paragraph{Synthetic report generation.}
\label{app:pseudo}
Because the four cohorts provide no radiology reports, we synthesize a report for each scan with the
same construction pipeline used for AbdomenAtlas 3.0~\cite{bassi2025radgpt}. Because these sets also
lack organ annotations, we first obtain multi-organ and suborgan masks with a pretrained MedFormer
segmentation model~\cite{bassi2025radgpt} and pair them with each set's ground-truth tumor masks. From the
combined masks a structured report is derived deterministically by the same construction as
RadGPT~\cite{bassi2025radgpt}, listing per suborgan the tumor count, size, location and attenuation,
as in \cref{sec:data}. OpenAI's \texttt{gpt-5.6-terra} then rewrites the structured report into a
narrative free-text report in the AbdomenAtlas 3.0 style, using the same system prompt as
RadGPT~\cite{bassi2025radgpt}.

\subsection{Additional ablations and analyses}

\subsubsection{LLM backbone ablation}
\label{app:mt-general}
\ours uses a $1$B medical pretrained LLM backbone. \Cref{tab:backbone} replaces it with the general-domain
InternVL3-2B and InternVL3-8B~\cite{zhu2025internvl3} under the identical multitask recipe and
segmentation module, on the test split as a 3-organ macro mean consistent with \cref{tab:main}. The
$1$B medical backbone attains higher DSC, instance-wise Recall, Precision and F1 than both general
backbones in both inference modes. Increasing the general backbone from 2B to 8B narrows the gap but
does not close it. This implies that domain-adapted language pretraining contributes more than
parameter count on this task, although the backbones also differ in family.
\begin{table}[!ht]\centering\footnotesize\setlength{\tabcolsep}{4pt}
\caption{LLM backbone ablation with the same recipe and segmentation module. Metrics as in
\cref{tab:main}, macro mean over three organs. \textbf{Bold}: highest per column.}\label{tab:backbone}
\resizebox{\columnwidth}{!}{%
\begin{tabular}{l cccc cccc}
\toprule
 & \multicolumn{4}{c}{\refseg} & \multicolumn{4}{c}{\gseg} \\
\cmidrule(lr){2-5}\cmidrule(lr){6-9}
Backbone & DSC & R & P & F1 & DSC & R & P & F1 \\
\midrule
InternVL3-2B (general) & 0.399 & 0.572 & 0.340 & 0.421 & 0.471 & 0.644 & 0.352 & 0.450 \\
InternVL3-8B (general) & 0.485 & 0.661 & 0.360 & 0.466 & 0.506 & 0.661 & 0.397 & 0.495 \\
\ours ($1$B medical)   & \textbf{0.510} & \textbf{0.674} & \textbf{0.397} & \textbf{0.499} & \textbf{0.523} & \textbf{0.677} & \textbf{0.433} & \textbf{0.528} \\
\bottomrule
\end{tabular}}
\end{table}

\subsubsection{Vision encoder initialization}
\label{app:encoder-init}
\ours pretrains its 3D CT encoder on two datasets, CT-RATE~\cite{hamamci2024ctrate} and
Merlin~\cite{blankemeier2024merlin}, to widen its body-region coverage. \Cref{tab:encinit} compares
this against initialization from the CT-RATE-only encoder~\cite{hamamci2024ctrate}, on the test split
with the same recipe. Our encoder yields
marginally higher recall and DSC, while precision is essentially equal.
\begin{table}[!ht]\centering\footnotesize\setlength{\tabcolsep}{4pt}
\caption{Vision encoder initialization with the same recipe, CT-RATE$+$Merlin for \ours vs.\ CT-RATE
only~\cite{hamamci2024ctrate}. Metrics as in \cref{tab:main}, macro mean over three organs.
\textbf{Bold}: highest per column.}\label{tab:encinit}
\resizebox{\columnwidth}{!}{%
\begin{tabular}{l cccc cccc}
\toprule
 & \multicolumn{4}{c}{\refseg} & \multicolumn{4}{c}{\gseg} \\
\cmidrule(lr){2-5}\cmidrule(lr){6-9}
Encoder init & DSC & R & P & F1 & DSC & R & P & F1 \\
\midrule
CT-RATE only          & 0.506 & 0.654 & \textbf{0.397} & 0.493 & 0.521 & 0.655 & 0.432 & 0.520 \\
\ours       & \textbf{0.510} & \textbf{0.674} & \textbf{0.397} & \textbf{0.499} & \textbf{0.523} & \textbf{0.677} & \textbf{0.433} & \textbf{0.528} \\
\bottomrule
\end{tabular}}
\end{table}

\subsubsection{Size-stratified segmentation metrics}
\label{app:size}
\Cref{tab:size} splits the \cref{tab:main} test-split tumor metrics at a tumor long-axis threshold of
$1$~cm. For every model the precision deficit is concentrated in the small ($<$1\,cm) band, whereas
the large ($\ge$1\,cm) band is comparatively well resolved. The promptable baselines lose almost all
precision on small tumors, with mean small-band \refseg precision $0.005$ for VoxTell, $0.019$ for
UniBioMed and $0.031$ for BiomedParse v2, against $0.179$ for \ours. BiomedParse v2 attains the
highest small-band recall at mean $0.521$ but the lowest large-band precision at $0.145$, indicating
dense over-segmentation rather than localization. The baseline precision remains low even on large
tumors, with mean $\ge$1\,cm \refseg precision $0.395$, $0.225$ and $0.145$ for VoxTell, UniBioMed and
BiomedParse v2, against $0.646$ for \ours. Report conditioning in \gseg raises \ours small-band recall
and precision over \refseg, lifting macro recall$_{<1}$ from $0.486$ to $0.582$ and precision$_{<1}$
from $0.179$ to $0.202$. Small-tumor precision remains low for all models, including \ours, for
example on pancreas it moves only from $0.013$ to $0.040$. For UniBioMed the report leaves the small
band essentially unchanged, with precision$_{<1}$ moving only from $0.019$ to $0.020$.
\begin{table}[!ht]\centering\footnotesize\setlength{\tabcolsep}{4pt}
\caption{\Cref{tab:main} metrics split at a $1$~cm tumor long axis; DSC$_{le}$: GT-referenced tumor
Dice. BiomedParse v2 has no \gseg mode. \textbf{Bold}: larger band per metric, organ and model.}\label{tab:size}
\resizebox{\columnwidth}{!}{%
\begin{tabular}{ll cccc cccc}
\toprule
 & & \multicolumn{4}{c}{Rs} & \multicolumn{4}{c}{Gs} \\
\cmidrule(lr){3-6}\cmidrule(lr){7-10}
Model & Size & DSC$_{le}$ & R & P & F1 & DSC$_{le}$ & R & P & F1 \\
\midrule
\multicolumn{10}{l}{\textbf{Liver}} \\
\multirow{2}{*}{VoxTell} & $<$1 & 0.152 & 0.272 & 0.007 & 0.013 & 0.108 & 0.191 & 0.003 & 0.006 \\
       & $\ge$1 & \textbf{0.384} & \textbf{0.559} & \textbf{0.400} & \textbf{0.466} & \textbf{0.339} & \textbf{0.499} & \textbf{0.337} & \textbf{0.402} \\
\multirow{2}{*}{UniBioMed} & $<$1 & 0.230 & 0.423 & 0.033 & 0.062 & 0.245 & 0.448 & 0.036 & 0.067 \\
       & $\ge$1 & \textbf{0.361} & \textbf{0.592} & \textbf{0.265} & \textbf{0.367} & \textbf{0.380} & \textbf{0.614} & \textbf{0.281} & \textbf{0.385} \\
\multirow{2}{*}{BiomedParse v2} & $<$1 & 0.319 & 0.571 & 0.048 & 0.089 & --- & --- & --- & --- \\
       & $\ge$1 & \textbf{0.515} & \textbf{0.763} & \textbf{0.160} & \textbf{0.264} & --- & --- & --- & --- \\
\multirow{2}{*}{\ours} & $<$1 & 0.268 & 0.475 & 0.226 & 0.306 & 0.275 & 0.486 & 0.264 & 0.342 \\
       & $\ge$1 & \textbf{0.466} & \textbf{0.696} & \textbf{0.585} & \textbf{0.636} & \textbf{0.459} & \textbf{0.682} & \textbf{0.644} & \textbf{0.662} \\
\midrule
\multicolumn{10}{l}{\textbf{Kidney}} \\
\multirow{2}{*}{VoxTell} & $<$1 & 0.105 & 0.182 & 0.006 & 0.012 & 0.064 & 0.113 & 0.004 & 0.008 \\
       & $\ge$1 & \textbf{0.529} & \textbf{0.711} & \textbf{0.459} & \textbf{0.558} & \textbf{0.417} & \textbf{0.572} & \textbf{0.541} & \textbf{0.556} \\
\multirow{2}{*}{UniBioMed} & $<$1 & 0.237 & 0.456 & 0.025 & 0.048 & 0.234 & 0.448 & 0.025 & 0.046 \\
       & $\ge$1 & \textbf{0.523} & \textbf{0.748} & \textbf{0.265} & \textbf{0.392} & \textbf{0.523} & \textbf{0.742} & \textbf{0.276} & \textbf{0.402} \\
\multirow{2}{*}{BiomedParse v2} & $<$1 & 0.370 & 0.706 & 0.040 & 0.076 & --- & --- & --- & --- \\
       & $\ge$1 & \textbf{0.713} & \textbf{0.939} & \textbf{0.156} & \textbf{0.268} & --- & --- & --- & --- \\
\multirow{2}{*}{\ours} & $<$1 & 0.351 & 0.625 & 0.298 & 0.403 & 0.349 & 0.615 & 0.303 & 0.406 \\
       & $\ge$1 & \textbf{0.681} & \textbf{0.904} & \textbf{0.794} & \textbf{0.845} & \textbf{0.684} & \textbf{0.902} & \textbf{0.822} & \textbf{0.860} \\
\midrule
\multicolumn{10}{l}{\textbf{Pancreas}} \\
\multirow{2}{*}{VoxTell} & $<$1 & 0.000 & 0.000 & 0.002 & 0.000 & 0.000 & 0.000 & 0.000 & 0.000 \\
       & $\ge$1 & \textbf{0.262} & \textbf{0.451} & \textbf{0.326} & \textbf{0.378} & \textbf{0.206} & \textbf{0.352} & \textbf{0.312} & \textbf{0.331} \\
\multirow{2}{*}{UniBioMed} & $<$1 & 0.035 & 0.143 & 0.000 & 0.000 & 0.047 & 0.214 & 0.000 & 0.001 \\
       & $\ge$1 & \textbf{0.181} & \textbf{0.412} & \textbf{0.145} & \textbf{0.214} & \textbf{0.187} & \textbf{0.422} & \textbf{0.166} & \textbf{0.238} \\
\multirow{2}{*}{BiomedParse v2} & $<$1 & 0.095 & 0.286 & 0.004 & 0.007 & --- & --- & --- & --- \\
       & $\ge$1 & \textbf{0.387} & \textbf{0.732} & \textbf{0.118} & \textbf{0.204} & --- & --- & --- & --- \\
\multirow{2}{*}{\ours} & $<$1 & 0.139 & 0.357 & 0.013 & 0.026 & 0.317 & 0.643 & 0.040 & 0.075 \\
       & $\ge$1 & \textbf{0.358} & \textbf{0.637} & \textbf{0.560} & \textbf{0.596} & \textbf{0.363} & \textbf{0.644} & \textbf{0.596} & \textbf{0.619} \\
\midrule
\multicolumn{10}{l}{\textbf{Mean}} \\
\multirow{2}{*}{VoxTell} & $<$1 & 0.086 & 0.151 & 0.005 & 0.008 & 0.057 & 0.101 & 0.002 & 0.005 \\
       & $\ge$1 & \textbf{0.392} & \textbf{0.574} & \textbf{0.395} & \textbf{0.467} & \textbf{0.321} & \textbf{0.474} & \textbf{0.397} & \textbf{0.430} \\
\multirow{2}{*}{UniBioMed} & $<$1 & 0.167 & 0.341 & 0.019 & 0.036 & 0.175 & 0.370 & 0.020 & 0.038 \\
       & $\ge$1 & \textbf{0.355} & \textbf{0.584} & \textbf{0.225} & \textbf{0.324} & \textbf{0.363} & \textbf{0.593} & \textbf{0.241} & \textbf{0.342} \\
\multirow{2}{*}{BiomedParse v2} & $<$1 & 0.261 & 0.521 & 0.031 & 0.057 & --- & --- & --- & --- \\
       & $\ge$1 & \textbf{0.538} & \textbf{0.811} & \textbf{0.145} & \textbf{0.245} & --- & --- & --- & --- \\
\multirow{2}{*}{\ours} & $<$1 & 0.253 & 0.486 & 0.179 & 0.245 & 0.314 & 0.582 & 0.202 & 0.274 \\
       & $\ge$1 & \textbf{0.502} & \textbf{0.746} & \textbf{0.646} & \textbf{0.692} & \textbf{0.502} & \textbf{0.743} & \textbf{0.687} & \textbf{0.714} \\
\bottomrule
\end{tabular}}
\end{table}

\subsubsection{Agentic inference pipeline: decomposing the source of false-positive suppression}
\label{app:reportgate}
In \cref{sec:exp-report}, the agentic inference pipeline supplies the report to \ours at inference
and is compared against the vision-only nnU-Net and report-supervised R-Super specialists. This
section decomposes that comparison, evaluating each model both on its own and under the agentic
inference pipeline to isolate how much of the false-positive suppression the pipeline contributes.
Without the pipeline, \refseg and \gseg each run a single whole-organ pass, issuing one tumor prompt
for the entire organ rather than decomposing it into the report-named suborgans.

\Cref{tab:reportgate} reports this decomposition on the in-domain test set. Even without the
pipeline, base \gseg exceeds the vision-only nnU-Net and R-Super specialists on liver and pancreas in
DSC, precision and F1, and base \refseg exceeds them in precision and F1, for example liver \gseg F1 $0.485$ against $0.286$ for nnU-Net and
$0.319$ for R-Super, while on kidney nnU-Net remains higher at F1 $0.553$. Base \gseg attains higher
F1 than base \refseg on all three organs, $0.485$, $0.495$ and $0.320$ against $0.438$, $0.453$ and
$0.249$. The agentic pipeline improves both further, chiefly through precision, raising liver \gseg
precision from $0.396$ to $0.559$ and F1 to $0.594$. With the pipeline the agentic \gseg exceeds both
pipeline-restricted specialists on liver and pancreas in F1, reaching $0.594$ and $0.475$ against the
best specialist F1 of $0.485$ and $0.452$, while nnU-Net attains the highest kidney F1 at $0.661$. 

\Cref{tab:appgating} repeats the decomposition on the four external cohorts. Each cohort has a single
GT-positive organ, so the base and $+$agentic rows differ only in whether the organ is prompted as a
whole or decomposed into the report-named suborgans. Averaged over the four cohorts, every variant of
\ours attains a higher precision and F1 than both specialists, with F1 from $0.470$ to $0.547$ against
$0.415$ for nnU-Net and $0.426$ for R-Super. \gseg$+$agentic also attains the highest mean DSC and
recall at $0.666$ and $0.790$. \refseg$+$agentic and both \gseg variants exceed the nnU-Net DSC of
$0.655$, which only base \refseg does not reach at $0.647$, and only \gseg$+$agentic exceeds the R-Super
recall of $0.786$.

Across both evaluation sets, the agentic pipeline benefits \refseg more than \gseg. On the in-domain
set, it raises mean \refseg F1 by $0.163$, from $0.380$ to $0.543$, and mean \gseg F1 by $0.115$, from
$0.433$ to $0.548$, mainly through precision, which rises from $0.271$ to $0.468$ for \refseg and from
$0.326$ to $0.478$ for \gseg. On the external cohorts, it raises mean \refseg F1 from $0.470$ to
$0.501$, mainly on CPTAC-CCRCC, where precision rises from $0.213$ to $0.303$. Mean \gseg F1 there is
comparable with and without the pipeline, at $0.547$ and $0.544$, as recall rises from $0.770$ to
$0.790$ and precision falls from $0.448$ to $0.432$. Most of this recall gain comes from the multifocal
CRLM cohort.

This pattern is consistent with the report already supplying \gseg the tumor locations that the
pipeline adds. The report names the organs and suborgans that contain tumors, whereas \refseg receives
only the referring phrase. The pipeline prompts only the report-named organs on the in-domain set and
decomposes each organ into its report-named suborgans on both sets, so its targets may add less new
information to \gseg than to \refseg. The suborgan targets do not carry the remaining report content, such
as attenuation, size and count. On the external cohorts, \refseg$+$agentic remains below base \gseg in
mean precision and F1, at $0.390$ and $0.501$ against $0.448$ and $0.547$. 
\Cref{app:relevancy} illustrates on three cases that \gseg places high relevancy on such report
attributes.
\begin{table}[t]\centering\footnotesize\setlength{\tabcolsep}{12pt}
\caption{In-domain decomposition of false-positive suppression. Base: \refseg/\gseg with one
whole-organ prompt, specialists unrestricted; $+$agentic: suborgan prompts, or for nnU-Net and R-Super,
predictions restricted to the reported organs. DSC: image-wise mean over GT-positive cases; R/P/F1 as in
\cref{tab:main}; Mean: macro mean over organs. \textbf{Bold}/\underline{underline}: best/second per
column.}\label{tab:reportgate}
\resizebox{\columnwidth}{!}{%
\begin{tabular}{ll cccc}
\toprule
Organ & Model & DSC & R & P & F1 \\
\midrule
\multirow{8}{*}{Liver}
 & nnU-Net & 0.603 & 0.568 & 0.191 & 0.286 \\
 & nnU-Net $+$agentic & 0.602 & 0.566 & 0.392 & 0.463 \\
 & R-Super & 0.580 & 0.518 & 0.230 & 0.319 \\
 & R-Super $+$agentic & 0.580 & 0.517 & 0.457 & 0.485 \\
 & \refseg & 0.600 & \underline{0.637} & 0.334 & 0.438 \\
 & \refseg $+$agentic & 0.610 & \textbf{0.659} & \underline{0.550} & \textbf{0.599} \\
 & \gseg & \textbf{0.629} & 0.624 & 0.396 & 0.485 \\
 & \gseg $+$agentic & \underline{0.612} & 0.633 & \textbf{0.559} & \underline{0.594} \\
\midrule
\multirow{8}{*}{Kidney}
 & nnU-Net & \textbf{0.677} & 0.765 & 0.433 & 0.553 \\
 & nnU-Net $+$agentic & \underline{0.664} & 0.751 & \textbf{0.590} & \textbf{0.661} \\
 & R-Super & 0.657 & \textbf{0.795} & 0.320 & 0.456 \\
 & R-Super $+$agentic & 0.652 & \underline{0.784} & 0.435 & 0.559 \\
 & \refseg & 0.612 & 0.765 & 0.321 & 0.453 \\
 & \refseg $+$agentic & 0.626 & 0.695 & \underline{0.503} & \underline{0.584} \\
 & \gseg & 0.629 & 0.758 & 0.367 & 0.495 \\
 & \gseg $+$agentic & 0.625 & 0.689 & 0.493 & 0.575 \\
\midrule
\multirow{8}{*}{Pancreas}
 & nnU-Net & 0.367 & 0.554 & 0.101 & 0.170 \\
 & nnU-Net $+$agentic & 0.367 & 0.554 & 0.297 & 0.387 \\
 & R-Super & 0.392 & 0.612 & 0.138 & 0.225 \\
 & R-Super $+$agentic & 0.392 & 0.612 & \underline{0.359} & \underline{0.452} \\
 & \refseg & 0.379 & 0.568 & 0.159 & 0.249 \\
 & \refseg $+$agentic & 0.393 & \underline{0.619} & 0.350 & 0.447 \\
 & \gseg & \textbf{0.406} & \textbf{0.626} & 0.215 & 0.320 \\
 & \gseg $+$agentic & \underline{0.403} & \textbf{0.626} & \textbf{0.383} & \textbf{0.475} \\
\midrule
\multirow{8}{*}{Mean}
 & nnU-Net & \underline{0.549} & 0.629 & 0.242 & 0.336 \\
 & nnU-Net $+$agentic & 0.544 & 0.624 & 0.426 & 0.504 \\
 & R-Super & 0.543 & 0.642 & 0.229 & 0.333 \\
 & R-Super $+$agentic & 0.542 & 0.638 & 0.417 & 0.499 \\
 & \refseg & 0.530 & 0.657 & 0.271 & 0.380 \\
 & \refseg $+$agentic & 0.543 & \underline{0.658} & \underline{0.468} & \underline{0.543} \\
 & \gseg & \textbf{0.554} & \textbf{0.669} & 0.326 & 0.433 \\
 & \gseg $+$agentic & 0.547 & 0.649 & \textbf{0.478} & \textbf{0.548} \\
\bottomrule
\end{tabular}}
\end{table}
\begin{table}[t]\centering\footnotesize\setlength{\tabcolsep}{5pt}
\caption{External-cohort version of \cref{tab:reportgate}; metrics as in \cref{tab:main}. One
GT-positive organ per cohort, so nnU-Net and R-Super appear once. Mean: macro mean over four cohorts.
\textbf{Bold}/\underline{underline}: best/second per column.}\label{tab:appgating}
\resizebox{\columnwidth}{!}{%
\begin{tabular}{ll cccc}
\toprule
Dataset & Model & DSC & R & P & F1 \\
\midrule
\multirow{6}{*}{HCC-TACE liver \scriptsize($n{=}130$)}
 & nnU-Net & 0.650 & 0.679 & 0.134 & 0.223 \\
 & R-Super & 0.597 & 0.653 & 0.111 & 0.190 \\
 & \refseg & 0.707 & 0.668 & 0.212 & 0.322 \\
 & \refseg $+$agentic & 0.719 & \underline{0.689} & 0.225 & 0.340 \\
 & \gseg & \underline{0.725} & \underline{0.689} & \textbf{0.302} & \textbf{0.420} \\
 & \gseg $+$agentic & \textbf{0.727} & \textbf{0.694} & \underline{0.293} & \underline{0.412} \\
\midrule
\multirow{6}{*}{CPTAC-PDA pancreas \scriptsize($n{=}108$)}
 & nnU-Net & 0.492 & 0.766 & 0.287 & 0.418 \\
 & R-Super & 0.471 & \underline{0.793} & \underline{0.449} & \underline{0.573} \\
 & \refseg & 0.475 & \textbf{0.802} & 0.387 & 0.522 \\
 & \refseg $+$agentic & 0.488 & \underline{0.793} & 0.389 & 0.522 \\
 & \gseg & \textbf{0.504} & \underline{0.793} & \textbf{0.476} & \textbf{0.595} \\
 & \gseg $+$agentic & \underline{0.503} & \textbf{0.802} & 0.443 & 0.571 \\
\midrule
\multirow{6}{*}{CPTAC-CCRCC kidney \scriptsize($n{=}108$)}
 & nnU-Net & 0.804 & \textbf{0.943} & 0.240 & 0.383 \\
 & R-Super & 0.791 & \textbf{0.943} & 0.187 & 0.312 \\
 & \refseg & 0.790 & \underline{0.935} & 0.213 & 0.346 \\
 & \refseg $+$agentic & 0.799 & 0.927 & 0.303 & 0.457 \\
 & \gseg & \textbf{0.808} & 0.927 & \underline{0.341} & \underline{0.499} \\
 & \gseg $+$agentic & \underline{0.807} & 0.927 & \textbf{0.348} & \textbf{0.506} \\
\midrule
\multirow{6}{*}{CRLM liver \scriptsize($n{=}197$)}
 & nnU-Net & \textbf{0.672} & 0.731 & 0.565 & 0.637 \\
 & R-Super & 0.611 & \textbf{0.756} & 0.539 & 0.630 \\
 & \refseg & 0.617 & 0.686 & \textbf{0.696} & \textbf{0.691} \\
 & \refseg $+$agentic & 0.625 & 0.731 & 0.643 & 0.684 \\
 & \gseg & 0.623 & 0.672 & \underline{0.674} & 0.673 \\
 & \gseg $+$agentic & \underline{0.626} & \underline{0.737} & 0.644 & \underline{0.687} \\
\midrule
\multirow{6}{*}{Mean}
 & nnU-Net & 0.655 & 0.780 & 0.307 & 0.415 \\
 & R-Super & 0.618 & \underline{0.786} & 0.322 & 0.426 \\
 & \refseg & 0.647 & 0.773 & 0.377 & 0.470 \\
 & \refseg $+$agentic & 0.658 & 0.785 & 0.390 & 0.501 \\
 & \gseg & \underline{0.665} & 0.770 & \textbf{0.448} & \textbf{0.547} \\
 & \gseg $+$agentic & \textbf{0.666} & \textbf{0.790} & \underline{0.432} & \underline{0.544} \\
\bottomrule
\end{tabular}}
\end{table}

\subsubsection{Report-word relevancy}
\label{app:relevancy}
\begin{figure}[!t]\centering
\includegraphics[width=0.92\columnwidth]{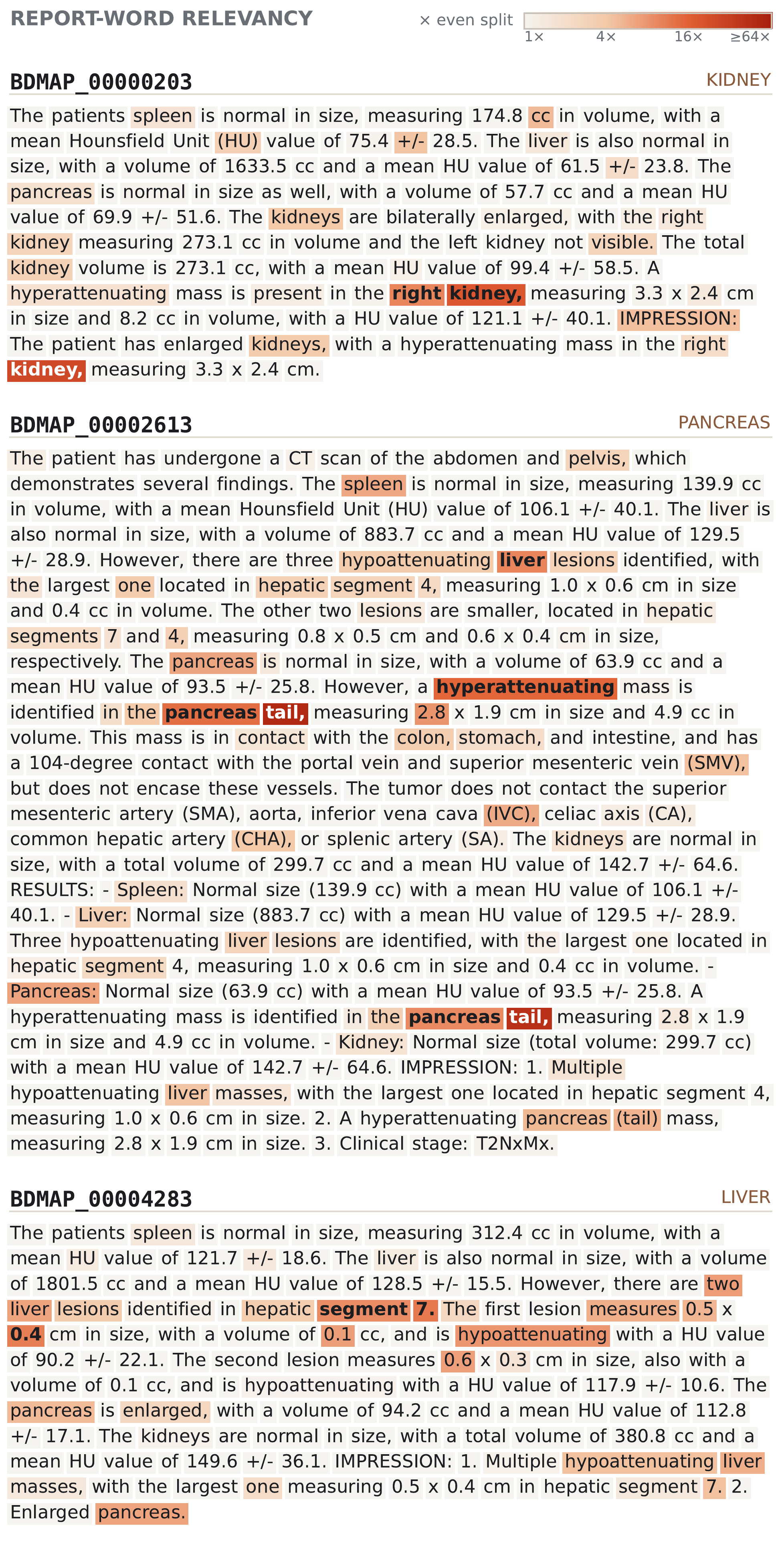}
\caption{Report-word relevancy for three in-domain \gseg cases, one per organ, by gradient-weighted
attention rollout~\cite{chefer2021generic}. Each word takes the maximum relevancy over its subword
tokens; color is its ratio to the mean word relevancy of its report, on a log scale shared across rows
from white at or below $1\times$ to $64\times$.}\label{fig:relevancy}
\end{figure}
We attribute the produced tumor mask to individual report words using the generic gradient-weighted
attention rollout of Chefer et al.~\cite{chefer2021generic}, computed over the language-model
self-attention. The backward target is the mean
foreground segmentation logit inside the reference tumor crop, so the attribution scores each report
token by its influence on the mask. Each word is scored by max-pooling the relevancy over its
subword tokens. We compute it on three in-domain
\gseg cases spanning kidney, pancreas and liver. Relative
to \refseg, the report clears $19$ false positives on the kidney case and raises DSC
from $0.285$ to $0.651$, moves the pancreas case from a missed to a recovered tumor at
DSC $0.097$ to $0.591$, and recovers a second true tumor on the multifocal liver case
at DSC $0.246$ to $0.436$.

\newcommand{\rrgok}[1]{\textcolor{green!45!black}{#1}}
\newcommand{\rrgbad}[1]{\textcolor{red!70!black}{#1}}
\begin{figure*}[t]\centering
\begin{minipage}[t]{\columnwidth}\centering
\includegraphics[width=0.55\columnwidth]{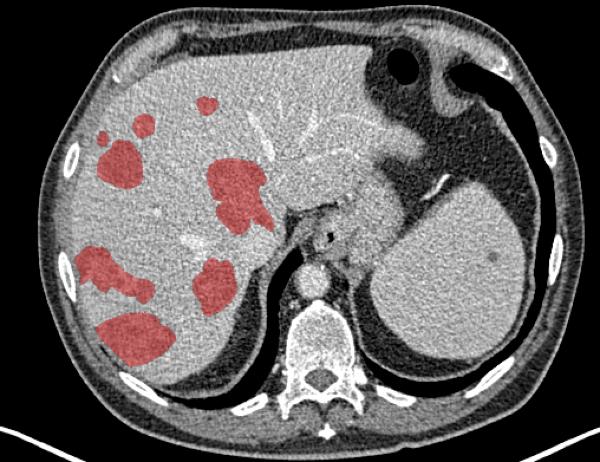}\par\smallskip
{\footnotesize\textbf{\VQA} (multifocal liver case)}\par\smallskip
\begin{userbubble}\textit{Yes/No.}~Are all liver lesions hypoattenuating?\end{userbubble}
\begin{modelbubble}\textbf{\ours:}~Yes\hfill{\scriptsize reference: Yes}\end{modelbubble}
\begin{ubmbubble}\textbf{UniBioMed:}~No\end{ubmbubble}
\begin{userbubble}\textit{Multiple choice.}~Attenuation of the largest liver lesion relative to liver?
(A)~hyperattenuating \; (B)~hypoattenuating \; (C)~isoattenuating \; (D)~heterogeneous\end{userbubble}
\begin{modelbubble}\textbf{\ours:}~B\hfill{\scriptsize reference: B}\end{modelbubble}
\begin{ubmbubble}\textbf{UniBioMed:}~T2b\end{ubmbubble}
\begin{userbubble}\textit{Short answer.}~Which hepatic segment contains the largest liver lesion?\end{userbubble}
\begin{modelbubble}\textbf{\ours:}~Segment 2\hfill{\scriptsize reference: Segment 7/8}\end{modelbubble}
\begin{ubmbubble}\textbf{UniBioMed:}~Posterior\end{ubmbubble}
\end{minipage}\hfill
\begin{minipage}[t]{\columnwidth}\centering
\includegraphics[width=0.55\columnwidth]{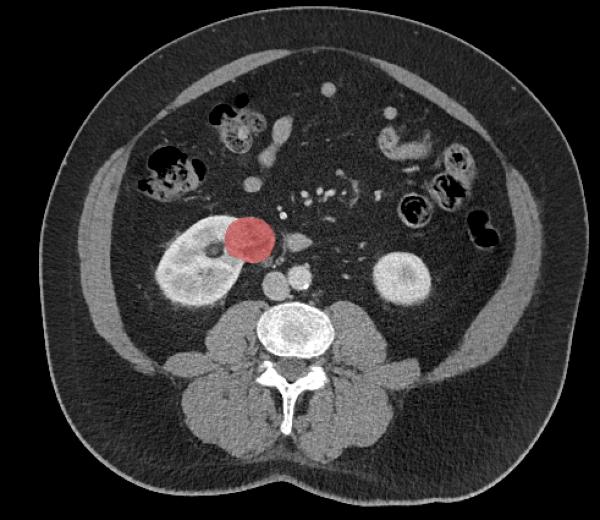}\par\smallskip
{\footnotesize\textbf{\RRG} (bilateral kidney case)}\par\smallskip
\begin{userbubble}\textit{Prompt.}~Generate the findings and impression for this CT.\end{userbubble}
\begin{modelbubble}\textbf{\ours:}~The patient has an \rrgok{enlarged spleen}, measuring 330.1\,cc in
volume. The liver is of \rrgok{normal size}, with a volume of 1853.1\,cc. The pancreas is \rrgok{also
enlarged}, with a volume of 103.1\,cc. The \rrgok{bilateral kidneys are enlarged}, with the right
kidney measuring 216.1\,cc and the left kidney measuring 221.2\,cc. There are \rrgbad{two} kidney
lesions identified. The first is located in the right kidney, measuring \rrgbad{1.0$\times$0.8\,cm},
and is \rrgok{hypoattenuating}. The second is located in the left kidney, measuring
\rrgbad{0.8$\times$0.6\,cm}, and is also \rrgok{hypoattenuating}. IMPRESSION: an \rrgok{enlarged spleen,
pancreas, and kidneys}, with \rrgbad{two} \rrgok{hypoattenuating kidney masses}, the largest in the
right kidney measuring \rrgbad{1.0$\times$0.8\,cm}.\end{modelbubble}
\begin{ubmbubble}\textbf{UniBioMed:}~-.\quad{\scriptsize\emph{(degenerate output, shown verbatim)}}\end{ubmbubble}
\begin{refbubble}\textbf{Reference:}~The patient has an enlarged spleen, measuring 349.8\,cc in volume.
The liver is of normal size. The pancreas is enlarged, with a volume of 155.2\,cc. Both kidneys are
also enlarged, with a total kidney volume of 493.3\,cc. Multiple hypoattenuating kidney lesions are
present. The largest is located in the right kidney, measuring 3.6$\times$3.3\,cm. Two smaller lesions
are present, one in the left kidney measuring 1.1$\times$0.9\,cm and another in the right kidney
measuring 1.1$\times$0.6\,cm. IMPRESSION: enlarged spleen, pancreas, and kidneys, with multiple
hypoattenuating kidney masses, the largest measuring 3.6$\times$3.3\,cm in the right
kidney.\end{refbubble}
\end{minipage}
\caption{\VQA on a multifocal liver case and \RRG on a bilateral-kidney case from the in-domain
cohort, each with one axial CT slice in radiological orientation and the reference tumor mask in red.
\rrgok{Green}/\rrgbad{red}: \ours expressions that match/contradict the reference, which is lightly
abridged.}\label{fig:qaexamples}
\end{figure*}
\Cref{fig:relevancy} shows the per-word relevancy over the narrative report for the three cases. Each
report states, per suborgan, the tumor count, size, location and attenuation, together with organ-level
descriptors such as organ volume and mean Hounsfield unit, and for the pancreas case an added
description of tumor contact with the abdominal vessels. We express each word score
as a ratio to an even split, namely the word score divided by the mean word score of its report. The
median word lies at $0.25\times$ to $0.33\times$ across the three cases. The highest ratio in each case
falls on the tumor location, namely \emph{kidney} in \emph{right kidney} at $25.5\times$ with
\emph{right} at $9.8\times$ on the kidney case, \emph{tail} at $50.2\times$ on the pancreas case, and the
segment number at $11.6\times$ with \emph{segment} at $8.9\times$ on the liver case. Attenuation
reaches $14.7\times$ for \emph{hyperattenuating} on the pancreas case and $7.8\times$ for
\emph{hypoattenuating} on the liver case, against $1.8\times$ on the kidney case. On the liver case the
count word \emph{two} reaches $7.2\times$ and the size figures up to $10.6\times$, and on the pancreas
case the size figure reaches $7.9\times$. On the pancreas case the vessel abbreviations \emph{IVC} and
\emph{SMV} reach $5.9\times$ and $4.4\times$, whereas the relation words \emph{contact} and
\emph{encase} reach $1.6\times$ and $0.5\times$. Organ names also reach
high ratios, including non-target organs such as \emph{liver} at $9.8\times$ and \emph{spleen} at
$6.2\times$ on the pancreas case. Organ-level measurements such as volume and mean Hounsfield unit stay
near $1.0\times$ on the pancreas case and below $5\times$ on all three cases. This indicates that
relevancy is highest on the report tumor location, followed by attenuation, size and count, and is at
most $1.6\times$ on the tumor-vessel relation. Because organ names without a tumor role also
reach high ratios, relevancy alone does not establish that a word is used.

\subsection{Language tasks}

\subsubsection{VQA and RRG}
\label{app:language}
In this section we report the language tasks on the in-domain test cohort, comparing \ours with the
UniBioMed baseline. For report generation UniBioMed conditions on sixteen randomly sampled 2D slices
rather than the full CT volume. We use exact-match accuracy per question type for VQA and BLEU,
ROUGE-L, METEOR~\cite{banerjee2005meteor} and BERTScore~\cite{zhang2020bertscore} for RRG. BERTScore-F1
uses DistilBERT (\texttt{distilbert-base-uncased}) following the radiology-report protocol of~\cite{yu2023evaluating}.
As shown in \cref{tab:language}, \ours exceeds the UniBioMed baseline on every metric. The gap is
largest on multiple-choice VQA, $0.640$ against $0.006$, and on report generation, where \ours attains
BERTScore-F1 $0.593$ against $0.073$ and ROUGE-L $0.377$ against $0.107$. Overall VQA accuracy is
$0.732$ against $0.294$. The report-generation gap may partly reflect this input bottleneck, since
the sampled 2D slices barely capture the quantitative content of these reports, namely organ volumes,
three-dimensional tumor sizes and tumor counts. In our runs the UniBioMed report head often returns
empty strings. \Cref{fig:qaexamples} gives one qualitative example for each task.

\begin{table}[!htb]\centering\footnotesize\setlength{\tabcolsep}{5pt}
\caption{Language tasks on the in-domain cohort, $n{=}373$ reports and $7{,}011$ VQA questions. MC:
multiple choice; B-1/B-4: BLEU-1/4; R-L: ROUGE-L; M: METEOR~\cite{banerjee2005meteor}; BS:
BERTScore-F1~\cite{zhang2020bertscore}. \textbf{Bold}: highest per column.}\label{tab:language}
\resizebox{\columnwidth}{!}{%
\begin{tabular}{l cccc ccccc}
\toprule
 & \multicolumn{4}{c}{VQA accuracy} & \multicolumn{5}{c}{RRG} \\
\cmidrule(lr){2-5}\cmidrule(lr){6-10}
Model & Y/N & MC & Short & All & B-1 & B-4 & R-L & M & BS \\
\midrule
UniBioMed & 0.434 & 0.006 & 0.004 & 0.294 & 0.072 & 0.006 & 0.107 & 0.086 & 0.073 \\
\ours     & \textbf{0.848} & \textbf{0.640} & \textbf{0.406} & \textbf{0.732} & \textbf{0.329} & \textbf{0.168} & \textbf{0.377} & \textbf{0.299} & \textbf{0.593} \\
\bottomrule
\end{tabular}}
\end{table}

\subsection{Computational cost}

\subsubsection{Inference latency}
\label{app:compute}
\begin{table}[!htb]\centering\footnotesize\setlength{\tabcolsep}{6pt}
\caption{Computational cost on one in-domain volume with $7$ suborgan targets, NVIDIA RTX A5000,
batch $1$, bfloat16, sliding-window pass only. Input: the original volume at $0.63\times0.63\times1.5$~mm,
which each model resamples to its own plan. Inference Time: mean over repeated runs, s.d.\ below
$0.1$\,s.}\label{tab:compute}
\resizebox{\columnwidth}{!}{%
\begin{tabular}{l cccc}
\toprule
Model & Params & Input ($W{\times}H{\times}D$) & Inference Time (s) \\
\midrule
nnU-Net (specialist) & $102$\,M  & $512\times394\times419$ & $6.2$ \\
R-Super (specialist) & $37.9$\,M & $512\times394\times419$ & $24.6$ \\
\midrule
\ours (\refseg) & $1.41$\,B & $512\times394\times419$ & $23.7$ \\
\ours (\gseg)   & $1.41$\,B & $512\times394\times419$ & $23.7$ \\
\bottomrule
\end{tabular}}
\end{table}
We compare inference cost against the two vision-only specialists, nnU-Net and R-Super. All models run
on the same GPU, an NVIDIA RTX A5000, on a single in-domain test volume. We time the network
sliding-window pass and average over repeated runs, with a per-value standard deviation below
$0.1$~s. The volume is the most tumor-involved case in the cohort, with $7$ suborgan targets across all
three organs, and approximates the upper end of \ours cost. Both specialists run a single vision-only
sliding-window pass. For \ours the measurement exercises its inference caching. The multimodal encoder
computes the image features once and memoizes them for reuse across all suborgan prompts. The report
prefix, namely the image tokens and, for \gseg, the narrative report, is prefilled once into the
language-model key-value cache. It is reused for every suborgan prompt, which differs only in the
trailing target phrase. The segmentation branch encodes the $192^3$ sliding window once per call and
shares it across targets, decoding one mask per suborgan from the common features. \refseg and \gseg
differ only in the prefill length. \Cref{tab:compute} shows that \refseg and \gseg have the same
measured inference time in this setting.

nnU-Net is fastest at $6.2$~s, and \ours and R-Super are comparable at $23.7$~s and $24.6$~s, with
R-Super timed under GPU-side window aggregation. R-Super and \ours run a $1$~mm isotropic sliding
window over the finer $324\times249\times628$ grid, with $128^3$ and $192^3$ patches. nnU-Net's
self-configured plan instead resamples to a coarser $0.86\times0.86\times2.0$~mm spacing, giving a
$376\times289\times314$ grid tiled with a $256\times192\times112$ patch, so it evaluates substantially
fewer windows. The \ours cost decomposes into a one-time encoding and a per-suborgan decode. The multimodal image encoding is negligible at $0.01$~s. The grounding
sliding-window encoding over the $1$~mm grid costs approximately $5$~s and is computed once per volume.
Each suborgan target then adds approximately $2.7$~s for its own mask decode, so latency grows linearly
with the number of targets, from $7.6$~s for one target to $23.7$~s for all seven.

\fi

\end{document}